\documentclass[10pt,twocolumn,letterpaper]{article}

\usepackage{cvpr}
\usepackage{times}
\usepackage{epsfig}
\usepackage{graphicx}
\usepackage{amsmath}
\usepackage{amssymb}
\usepackage{verbatim}
\usepackage{caption}
\usepackage{booktabs}
\usepackage{tabularx}
\usepackage{array}
\usepackage{multirow}
\usepackage{subcaption}

\usepackage[pagebackref=true,breaklinks=true,letterpaper=true,colorlinks,bookmarks=false]{hyperref}

\cvprfinalcopy 

\def\cvprPaperID{4317} 

\ifcvprfinal\fi
\begin{document}

\title{Birds of a Feather: Capturing Avian Shape Models from Images}


\author{Yufu Wang 
\and Nikos Kolotouros
\and Kostas Daniilidis
\and Marc Badger 
\and University of Pennsylvania\\
{\tt\small \{yufu, nkolot, kostas, mbadger\}@seas.upenn.edu}\\
}

\twocolumn[{%
\renewcommand\twocolumn[1][]{#1}%
\maketitle
\begin{center}
  \newcommand{\teaserwidth}{0.98\textwidth}
 \vspace{-0.3in}
    \centerline{
    \includegraphics[width=\teaserwidth,clip]{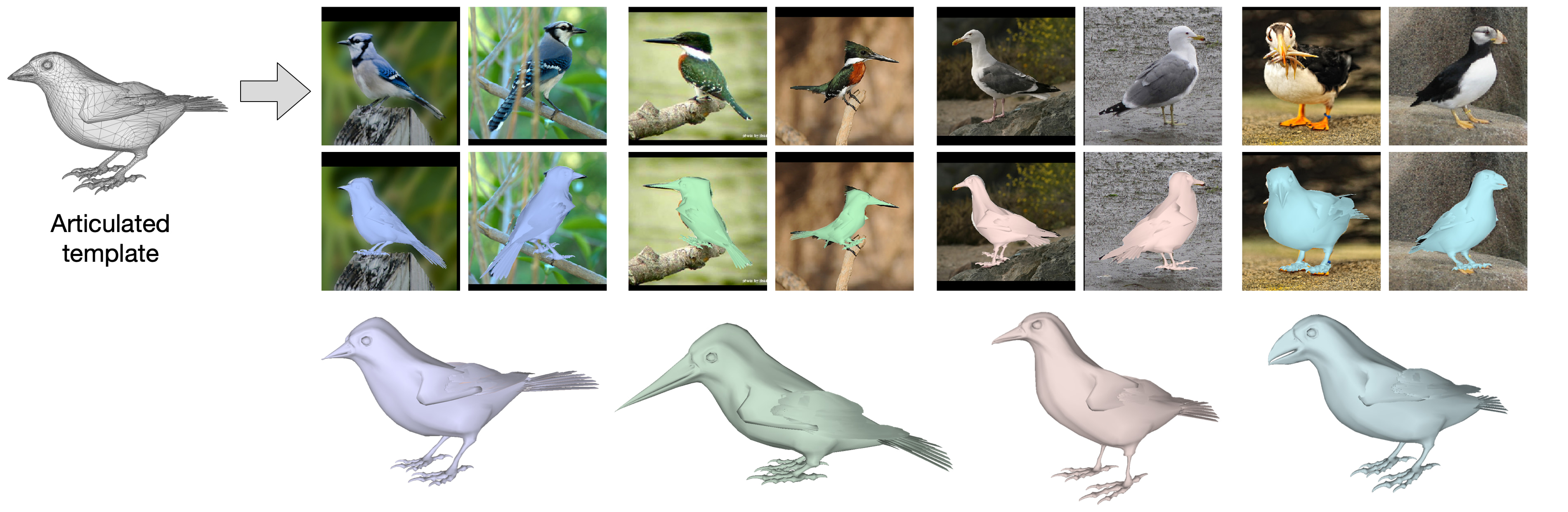}
    }
    \captionof{figure}{\textbf{Capturing shape models from images}. 
    We deform an articulated template to capture species-specific shape models from CUB image collections~\cite{wah2011caltech}. The new shape models not only articulate but can also deform according to species-specific shape deformation modes. We combine models from diverse species to learn a multi-species model.
}

\label{fig:teaser}
\end{center}}]


\begin{abstract}
   Animals are diverse in shape, but building a deformable shape model for a new species is not always possible due to the lack of 3D data. We present a method to capture new species using an articulated template and images of that species. In this work, we focus mainly on birds. Although birds represent almost twice the number of species as mammals, no accurate shape model is available. To capture a novel species, we first fit the articulated template to each training sample. By disentangling pose and shape, we learn a shape space that captures variation both among species and within each species from image evidence. We learn models of multiple species from the CUB dataset, and contribute new species-specific and multi-species shape models that are useful for downstream reconstruction tasks. Using a low-dimensional embedding, we show that our learned 3D shape space better reflects the phylogenetic relationships among birds than learned perceptual features. Project website: \tt\small\color{magenta} https://yufu-wang.github.io/aves/
   
   
\end{abstract}

\section{Introduction}

Automated capture of animal shape and motion is a challenging problem with widespread application in agriculture, biomechanics, animal behavior, and neuroscience. Changes in shape can convey an animal's health status and transmit social signals~\cite{alello2016vet, morris1956social}.  Recent methods that use articulated mesh models to extract these signals from images~\cite{badger20203d, biggs2020left, zuffi2019three, zuffi2018lions} are poised to transform these fields. One challenge that prevents wider adoption of this approach, however, is the difficulty of obtaining suitable models for new species. Methods that can automatically capture the shape of new species are highly desirable.

Recent articulated, 3D animal shape models obtain training data from 3D scans of toy animals figurines, or multiple views of the same subject~\cite{badger20203d, zuffi20173d, zuffi2018lions}. They produce great quality of reconstruction when the target is represented in the source data. How these articulated and deformable models can be extended to new animal species is still an open problem, particularly due to the lack 3D scans. Animals' non-cooperative nature makes obtaining 3D scans impractical.

Images of the same species, on the other hand, are more readily available for a wider variety of categories. The CUB~\cite{wah2011caltech} dataset, for example, provides image collections for various bird species. These monocular collections remain underutilized, however, because current detailed capture methods require a strong deformable 3D shape prior, which is not available for many species.

We propose a method to directly capture articulated shapes from species-specific image collections, when a strong deformable shape prior is not available. We focus our effort on CUB and start with an articulated bird mesh~\cite{badger20203d} as a generic template model. For a given collection, we first align the mesh to each annotated instance by solving for the pose and camera parameters. We then update the template model through a series of deformations to better fit the silhouettes, resulting in recovery of a new species-specific shape and individual variations within the species.

Closest to our work is SMALR~\cite{zuffi2018lions}, which uses a deformable shape model and a video sequence of the same instance to reconstruct quadrupeds. We relax this assumption and use a collection of different instances. This breaks the same-subject constraints and makes a naive adaption of SMALR infeasible. Solving this ``multiple instances'' problem allows us to build species-specific morphable shape models directly from images. Our method also starts with a simpler shape prior, using an articulated mesh as the template model.

To handle these challenges, we explicitly model two levels of the shape hierarchy. The first level is the difference between the generic template model and the average shape of a new species; the second level is the variation among individuals within that species.

First, after aligning the articulated mesh to the images for a novel species, we optimize a per-vertex deformation to bridge the difference between the shape of the template model and the shape of a novel species. We call the resulting shape the species mean shape. In the second step, starting from the mean shape, we learn the variation within the collection as a blend shape basis, allowing us to reconstruct each sample as a combination of shape basis vectors on top of the estimated mean. The shape basis also provides a species-specific morphable shape model. Because the articulation is factored out during model alignment, the mean and shape basis are properly captured without the nonlinear effect of pose.

Additionally, from all the per-species models we learn a new parametric shape model, AVES, that is capable of representing multiple avian species. We demonstrate through experiments that AVES captures a meaningful shape space, generalizes to unseen samples, and can be integrated in a deep learning pipeline.

In summary, our contributions are the following:
\begin{itemize}
    \item We present a new method that recovers detailed, species-specific shape models for a range of bird species using an articulated 3D mesh and  images of different instances of each species.
    \item We provide AVES, a multi-species statistical shape model extracted from reconstructions of 17 bird species. We show that the AVES shape space captures morphological traits that are correlated with the avian phylogeny.
    \item We show that the AVES model can be used in downstream reconstruction tasks including model fitting and regression, outperforming previous model-based and model-free approaches by a large margin.
\end{itemize}

\section{Related Work}
Our goal is to capture articulated 3D animal shapes from species-specific images. Here we focus on approaches that are most relevant to our problem, and review how they have been applied to humans and animals. 

\textbf{Model-based Reconstruction.} The reconstruction of non-rigid and articulated objects benefits greatly from a strong prior. A wealth of methods employ parametric models and treat the reconstruction problem as a parameter estimation problem. For human body, such models are learned from thousands of registered 3D scans~\cite{allen2003space, anguelov2005scape, joo2018total, loper2015smpl, osman2020, pavlakos2019expressive, xu2020ghum}; SMPL~\cite{loper2015smpl} being the most widely used.

For animals where 3D scans are impractical, various adaptations have been proposed. Zuffi \etal~\cite{zuffi20173d} introduced SMAL, an articulate quadruped model parameterized similar to SMPL and learned from 3D scans of toy figurines. Biggs \etal~\cite{biggs2020left} extend the SMAL model to include limb scaling to model dog breeds. Badger \etal~\cite{badger20203d} similarly parametrized an articulated bird mesh and used multi-view data from an aviary to provide pose and shape priors. 

These models can be fitted to different modality of sensor data or annotations in an optimization paradigm~\cite{bogo2016keep, huang2017towards, weiss2011home, zuffi20173d}. Deep learning has  made directly regressing model parameters possible~\cite{kanazawa2018end,kolotouros2019convolutional, kolotouros2019learning, omran2018neural, pavlakos2018learning}. Similarly, these techniques are adapted to regress parameters for articulated animal models when training data is available~\cite{badger20203d, biggs2020left}.

\textbf{Recovery beyond Parametric Models.}
Building details on top of a parametric model has the advantage that shape and pose information are decoupled and the recovered shape can be easily re-animated. Alldieck \etal~\cite{alldieck2018detailed, alldieck2018video} use video sequences of the same human to optimize a non-rigid deformation on SMPL to reconstruct details in hair, clothing and facial structure. Octopus~\cite{alldieck2019learning} then obtains comparable results by training a network with synthetic data to predict the deformation using multiple views and test-time optimization-based refinement. ARCH~\cite{huang2020arch} learns to estimate an implicit surface around a rigged body template to recover detailed and animatable avatars from a single image.

Similarly for animals, Zuffi \etal~\cite{zuffi2018lions} non-rigidly deforms SMAL with video sequences to capture detailed and textured quadrupeds. Three-D Safari~\cite{zuffi2019three} 
uses this method to capture 10 realistic zebras to generate synthetic data and train a network to predict zebra pose and shape.

Our approach also captures deviation from a parametric model. Differently, we relax the requirement of having a single subject in video sequence, and instead capture deformations in a ``multiple instance" setting.

\textbf{Model-free Capturing.}
There is a large amount of works on human shape capture without a parametric model but they tend to be supervision-heavy and do not immediately generalize to animals. Please refer to \cite{habermann2020deepcap} for a more comprehensive review. 

For animals, ``model-free" methods focus on lowering the requirement of supervision or prior knowledge. Early methods employ inflation techniques to extract shapes from silhouettes~\cite{ntouskos2015component, vicente2013balloon}. More recent methods learn end-to-end predictors, such as CMR~\cite{kanazawa2018learning}, U-CMR~\cite{goel2020shape} and IMR~\cite{tulsiani2020implicit}, to produce textured 3D animal mesh from a single image. So far, the outputs of these methods have limited realism and cannot be re-animated.

\textbf{Learning Shape Models.}
We aim to capture the shape space of a new species in the form of a morphable basis model. Shapes are classically defined over the geometry that is invariant only to Euclidean similarity transform~\cite{dryden1998statistical}. Many traditional deformable models treat the same object (e.g. hands) going under articulation to have different shapes~\cite{bregler2000recovering, cootes1995active, heap1996towards, zhou2016sparse}. As a result, their basis encode both pose and shape.

More recent methods often see benefits in disentangling pose and shape for articulated objects~\cite{anguelov2005scape, li2017learning, loper2015smpl, romero2017embodied, zuffi20173d}. 
Their shape space captures deformation intrinsic to identity, or animal species. Unfortunately, most of these models are learned from registered 3D data that are generally not available for many animal categories. Other methods have also learned deformable models from images, but they either do not disentangle articulation and shape, or only apply to rigid categories~\cite{cashman2012shape, kanazawa2018learning, kar2015category, torresani2008nonrigid}. Our approach uses an articulated mesh to explicitly separate pose before learning a meaningful basis shape space directly from annotated images.



\section{Approach}

We use the recent parametric bird model~\cite{badger20203d} as a starting point. For each sample in a species' image collection, we first align the model, through articulation and translation, to the annotated silhouette and keypoints. We then fix the alignment and update the shape in order to improve reconstruction on each image. We decompose this update into two steps. In the first step we optimize a per-vertex deformation on the shape template that is shared for every sample. This step transforms the generic template shape to an estimated mean for the new species. Finally, on top of this mean shape, we optimize a set of shape basis vectors and their coefficients for each sample to capture variation within the collection.


\begin{figure*}[hbt!]
\begin{center}
\includegraphics[width=0.98\linewidth]{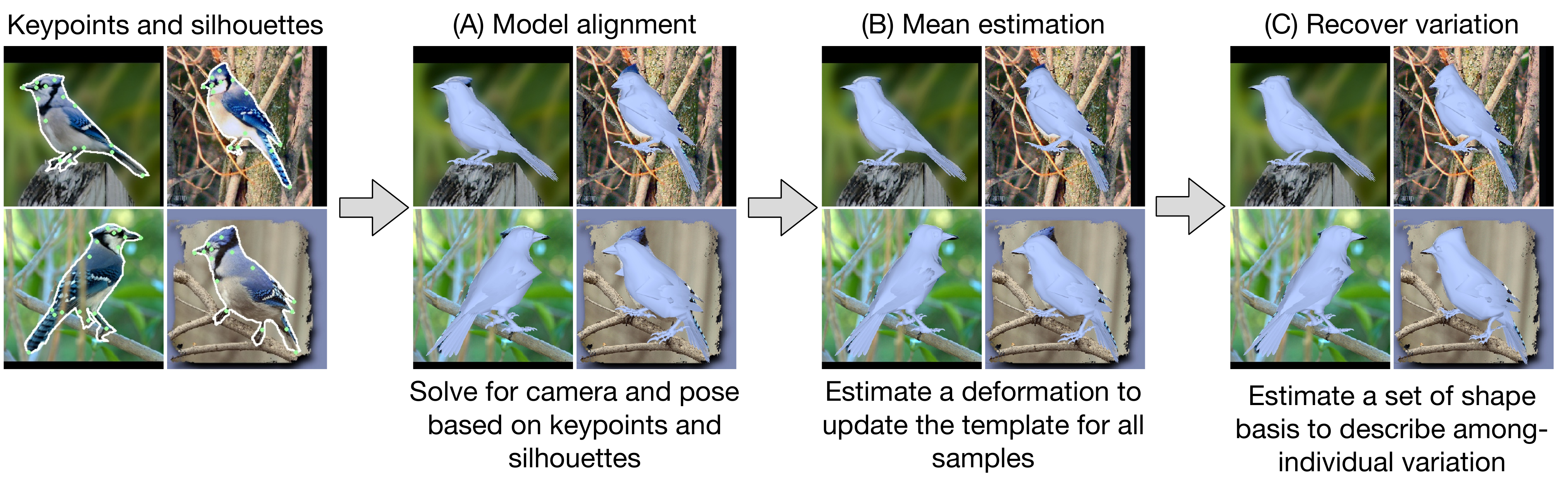}
\end{center}
\vspace{-0.25in}
\caption{\textbf{Method description}. We align an articulated template to images of a given species (in this case, blue jay), and deform it to first capture the species mean shape (B) and subsequently the shape variations across individuals (C). The mean shape deformations are the same among individuals in the same class, whereas the identity-specific offsets are expressed as linear combinations of a learned blend shape basis.}
\vspace{-0.1in}
\label{fig:method}
\end{figure*}

\subsection{Articulated bird model}
The articulated bird model (ABM)~\cite{badger20203d} is a function $M(\theta, \alpha, \gamma)$ of pose $\theta$, bone length $\alpha$, and translation $\gamma$. Pose $\theta \in \mathbb{R}^{3J}$ is the axis-angle representation of each joint orientation in the kinematic skeleton. The bone parameter $\alpha \in \mathbb{R}^{J}$ scales the distance between neighbouring joints. It allows body proportion to slightly vary and additionally models non-rigid joint movements that are common in birds, \eg the stretching of a bird's neck. Finally, $\gamma \in \mathbb{R}^{3}$ applies a translation to the root joint. The function $M(\theta, \alpha, \gamma)$ then articulates a template mesh $\textbf{v}_{bird} \in \mathbb{R}^{3N}$ through linear blend skinning and returns a 3D mesh.

We aim to align ABM to different species, but accurate alignment is more difficult when the new species has very different beak or tail length. Similar to \cite{biggs2020left}, we augment the model with two local scaling parameters, $\kappa$, that scale the length of the beak and tail respectively. The resulting model becomes $M(\theta, \alpha, \gamma, \kappa)$. Though such scaling is not always realistic, it can be refined during the deformation steps.

\subsection{Alignment to images}
To align the articulated mesh to images, we adopt a regression followed by optimization approach similar to \cite{badger20203d, zuffi2019three}. First we use a regression network that takes 2D keypoint locations as input and predicts model parameters: $\alpha, \theta, \gamma$. This is then used as initialization for an optimization procedure that refines the alignment. 

The regression network consists of two fully connected layers, each followed by batch normalization and max-pooling, and a final layer that predicts all model parameters. Different than \cite{badger20203d}, we do not include the silhouette as input to predict $\alpha$; because the bone parameter $\alpha$ mainly captures body proportion changes and this information can be inferred from keypoints alone. We train the network with synthetic keypoints-parameters pairs that are generated by animating the model with its pose priors.

Using the regression result as initialization, we minimize keypoints and silhouette reprojection error and a pose prior regularization with respect to $\Theta^{(i)} = \{\alpha, \theta, \gamma, \kappa\}$ for each image $i$ independently. We  omit notation $(i)$ in this step. Specifically, we define the objective as:
 \begin{equation}
E(\Theta) = E_{kp}(\Theta) + E_{msk}(\Theta) + E_{prior}(\alpha, \theta)
\label{eq:obj_pose}
\end{equation}

\textbf{Keypoint reprojection.} The keypoint reprojection term penalizes the distances between annotated keypoint locations and reprojection of the mesh keypoints. Denoting $P_k(\Theta)$ as a function that returns the $k^{th}$ keypoints from the articulated mesh, $\Pi = \Pi(P_k(\Theta))$ as its projection, $\Pi(x)$ as the camera model, and $\mathcal{P}_{k}$ as the ground truth, the keypoint term can be expressed as 
\begin{equation}
E_{kp}(\Theta) = \sum_{\mathrm{kpt} \, k}\rho(\parallel\Pi(P_k(\Theta) - \mathcal{P}_{k}\parallel_2)
\label{eq:keypoint_loss}
\end{equation}
where $\rho$ is the robust Geman-McClure function~\cite{geman1987statistical}. We use the perspective camera model for $\Pi(x)$ and a fixed focal length.

\textbf{Silhouette reprojection.} The silhouette term penalizes the discrepancy between the ground truth mask and the reprojected mask. Here we denote $\mathcal{R}$ as the differentiable rendering of the mesh~\cite{ravi2020pytorch3d}, and $\mathcal{S}$ as the ground truth. Then the silhouette term is
\begin{equation}
E_{msk}(\Theta) = \lambda_{msk} L_{\delta}(\mathcal{R}(M(\Theta)) - \mathcal{S}).
\label{eq:mask_loss}
\end{equation}
where $\lambda_{msk}$ is the importance weighting, and $L_{\delta}$ is the smooth $L1$ loss~\cite{girshick2015fast}. 

\textbf{Prior regularization.} We use the means and covariance of $\alpha$ and $\theta$ provided by the model, and define the prior term as the squared Mahalanobis distance similar to previous approaches~\cite{badger20203d,zuffi20173d,zuffi2018lions} to regularize the optimization.

Figure \ref{fig:method} (A) illustrates the resulting alignments.

\subsection{Obtaining species-specific shape}
After aligning ABM to each image, we perform the first step of the shape update to better explain the image cues. We define the deformation as a per-vertex displacement vector $\textbf{dv} \in \mathbb{R}^{3N}$ that updates the template shape as
\begin{equation}
\textbf{v}_{shape} = \textbf{v}_{bird} + \textbf{dv}
\label{eq:mean_shape}
\end{equation}
before different articulations are applied. This deformation vector, shared across all samples, can be seen as a transformation on the original generic shape, to bring it closer to that of the new species. 

The species-specific image collection provides guidance with silhouettes and keypoints; $\textbf{dv}$ will have to explain the discrepancy between the template and some new features presented by the new species. What we hope to achieve is to offset the template to a shape better suited for the new species, which will also condition the next stage when we reconstruct more details for each sample. We  call this intermediate shape the species mean shape.

To find the species mean, we fix model parameter $\Theta$ and minimize the following objective with respect to $\textbf{dv}$, 
 \begin{equation}
E(\textbf{dv}) = \sum_i E_{kp}^{(i)}(\textbf{dv}) + E_{msk}^{(i)}(\textbf{dv}) + E_{sm}(\textbf{dv})
\label{eq:obj_mean_pose}
\end{equation}
where $E_{kp}$ and $E_{msk}$ are the keypoint and silhouette reprojection terms from \textbf{3.2}, but now influenced by the changing shape $\textbf{v}_{bird}$ + \textbf{dv} and are summed over all instances in the collection.

We implement the smoothing term $E_{sm}(\textbf{dv})$ as
 \begin{equation}
E_{sm}(\textbf{dv}) = E_{edge} + E_{lap} + E_{arap} + E_{sym}
\label{eq:obj_mean_pose}
\end{equation}
$E_{edge}$ smooths displacements between adjacent vertices and is defined as $E_{edge}=\sum_{(p,q) \in edges} \parallel \textbf{dv}_p - \textbf{dv}_q \parallel_2$. $E_{lap}$ performs Laplacian smoothing on \textbf{dv}~\cite{nealen2006laplacian}. To preserve local details, $E_{arap}$ enforces as-rigid-as-possible regularization on $\textbf{v}_{shape}$~\cite{sorkine2007rigid}. The leg and claw regions are given a higher rigidity. Finally, $E_{sym}$ encourages the overall shape to be symmetrical~\cite{zuffi2018lions}. A weighted combination of these terms is used to produce the best outcome.

After minimizing the objective with respect to \textbf{dv}, we arrived at an estimated mean, as shown in Figure \ref{fig:method} (B), that explains image evidence better than the generic shape but still needs refinement to fit each individual.

\subsection{Reconstructing individuals}
Starting from the estimated mean, we aim to reconstruct subject-specific details for each sample. This problem is poorly constrained as we only have one view per subject. However, we can assume that their shapes are drawn from the same distribution. 
If the estimated mean shape from last step is a good approximation of the species mean, we can model the shape variation around the mean with a set of basis vectors.

For each individual $i$ in the collection, its shape can be updated from the last step as 
\begin{equation}
\textbf{v}_{shape}^{(i)} = \textbf{v}_{bird} + \textbf{dv} + \sum_{j}^{K}\beta_j^{(i)}*\textbf{dv}_j
\label{eq:individual_shape}
\end{equation}
where $\{\textbf{dv}_j\}$ is the set of $K$ basis vectors that describe the variation around the species mean, and $\{\beta_j^{(i)}\}$ are the coefficients for individual $i$. We can rewrite $\sum_{j}^{K}\beta_j^{(i)}*\textbf{dv}_j$ as $\textbf{V}\beta^{(i)}$ where $\textbf{V}$ is the matrix of basis vectors and $\beta^{(i)}$ is the coefficient vector for sample $i$.

To find $\textbf{V}$ and $\beta^{(i)}$, we minimize the following objective:
\begin{equation}
E(\textbf{V}, \beta^{(i)}) = \sum_i [E_{kp}^{(i)}
+ E_{msk}^{(i)}
+ E_{sm}^{(i)}](\textbf{V}, \beta^{(i)})
\label{eq:obj_individual_shape}
\end{equation}
where $E_{kp}$, $E_{msk}$ and $E_{sm}$ are defined similarly as in \textbf{3.3} but now a function of $\textbf{V}$ and $\beta^{(i)}$. We also experiment with a soft orthogonality constraint as $\parallel \textbf{V}^{T}\textbf{V} - I \parallel_F$ to capture uncorrelated features, and to enforce the magnitude of each basis vector to be constant to resolve the scale ambiguity between the basis and the coefficients. But empirically we do not find improvement in the results.

We obtain a detailed shape for each sample after minimizing the objective. Figure \ref{fig:method} (C) shows results for reconstructing blue jay images (4 of 12 shown).

After reconstruction, we re-learn the mean and the shape basis via PCA to arrive at the final species-specific deformable shape model.

\begin{figure}
\begin{center}
\includegraphics[width=0.95\linewidth]{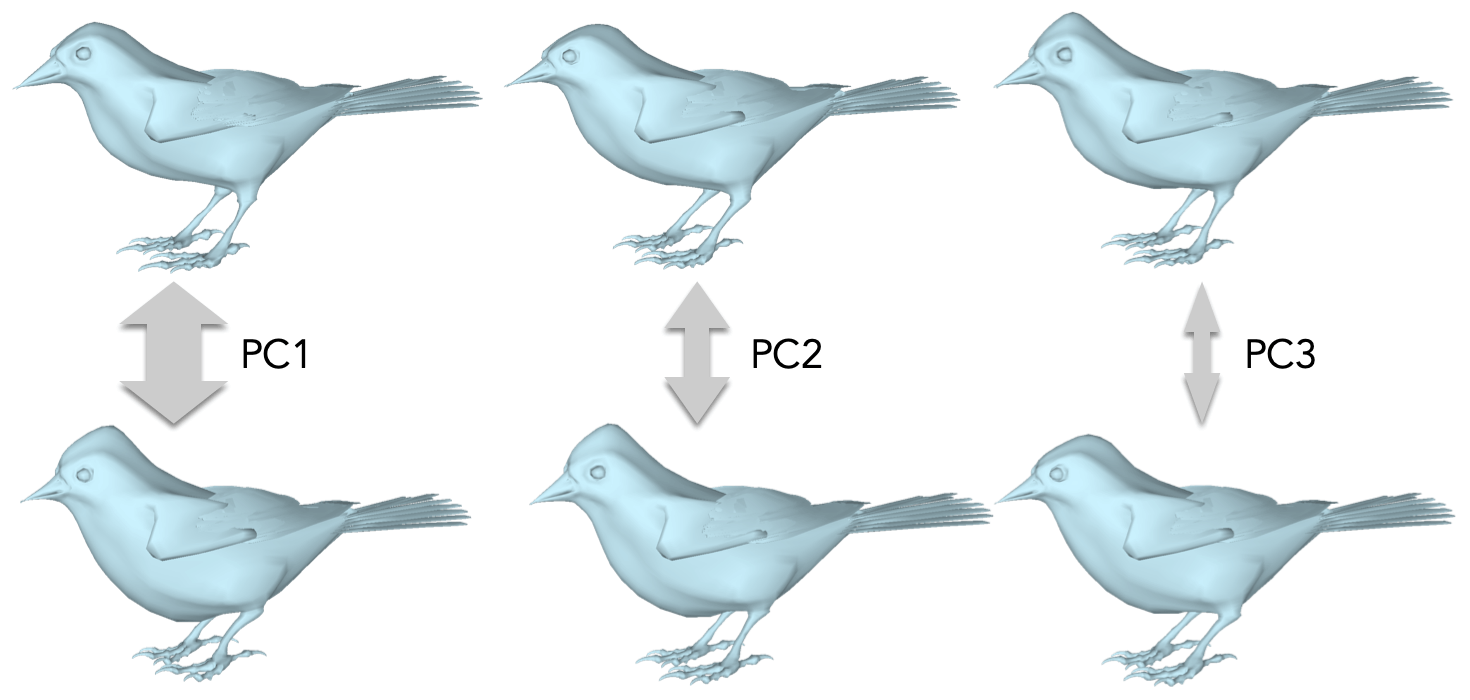}
\end{center}
   \caption{\textbf{Principle directions for the blue jay model}, shown with $\pm$1.5std. After learning, we optionally subdivided the surface of the learned models once for higher smoothness; same in Figure \ref{fig:ss_all}. Comparison in Sup.Mat.}
\label{fig:ss_bluejay}
\end{figure}

\begin{figure}
\begin{center}
\includegraphics[width=0.95\linewidth]{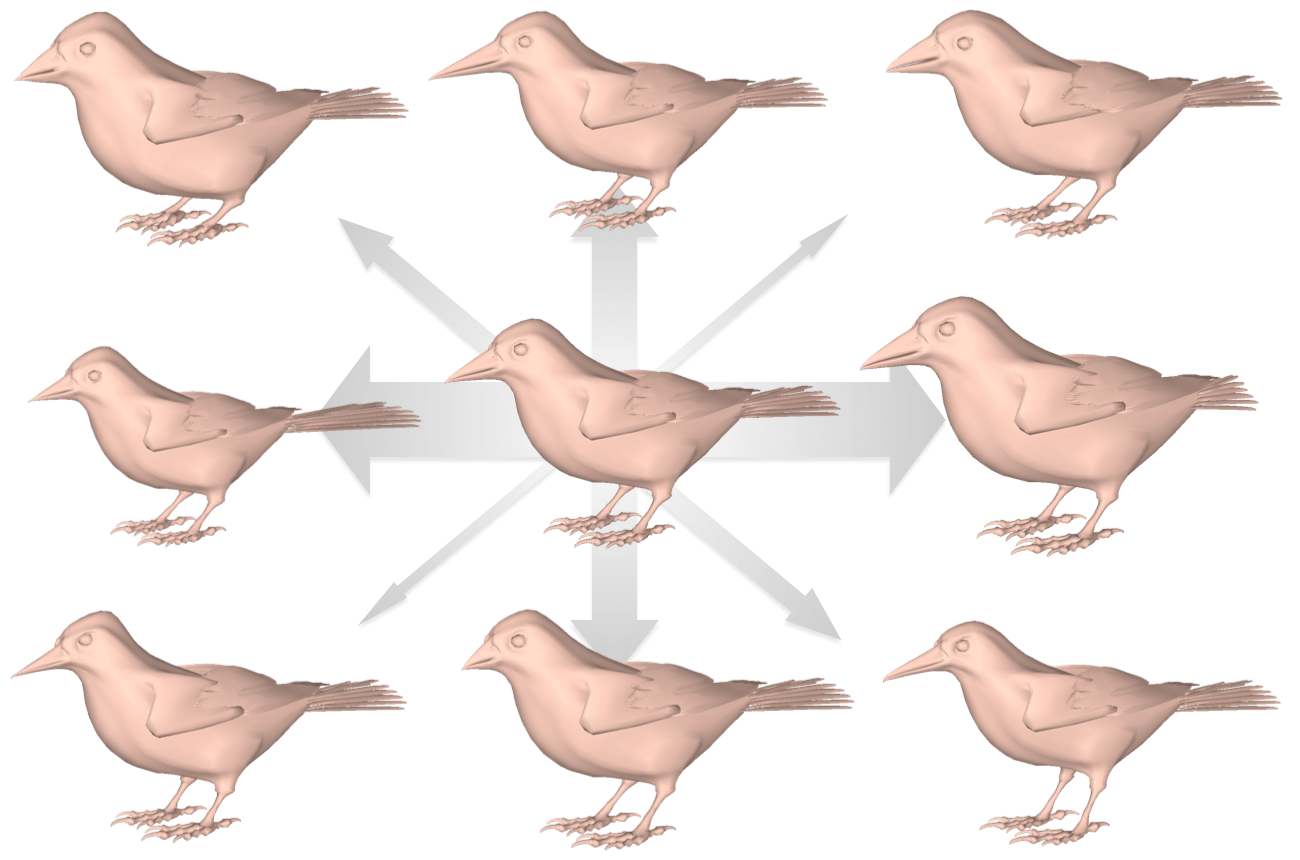}
\end{center}
   \caption{\textbf{The first 4 principle directions in the multi-species shape space}. The first two directions are shown with $\pm$1std and the rest with $\pm$2std. Arrow width from thick to thin indicates the first through fourth principle directions, respectively.}
\label{fig:ss_all}
\end{figure}

\section{Experiments}
In this section we empirically show that our method is able to capture realistic shapes and that the recovered shape space is meaningful from a biological standpoint. We also evaluate the accuracy of the methods and the generalization of the learned models. Additionally we incorporate our parametric model in a regression framework that outperforms previous 3D reconstruction methods.

We learn the parametric models using the CUB-200 dataset~\cite{wah2011caltech}. CUB contains 200 different bird species with segmentation and keypoint annotations. Certain keypoints are not useful for reconstruction, so we adapt only 8 of the original keypoints and additionally hand annotate 10 new keypoints for each sample in our experiments. We select 17 species representing a diverse shape collection. Since the main goal is to accurately capture shapes, we follow previous practice~\cite{alldieck2019learning, alldieck2018detailed, zuffi2018lions} and avoid samples whose pose is difficult to model, including flying and heavily occluded samples. Moreover, if a sample fails the alignment step, we do not include it in the learning steps. On average, we used 18 samples to learn each species. Results in Figure \ref{fig:reconstruction} shows realistic captures of different species. More results are included in the supplementary material. 

We use blue jays as an example and show variation captured by the species-specific shape basis in Figure \ref{fig:ss_bluejay}. We observe variation among individuals in body type, crown size, crown direction and chest. 

\begin{figure}
\begin{center}
\includegraphics[width=0.98\linewidth]{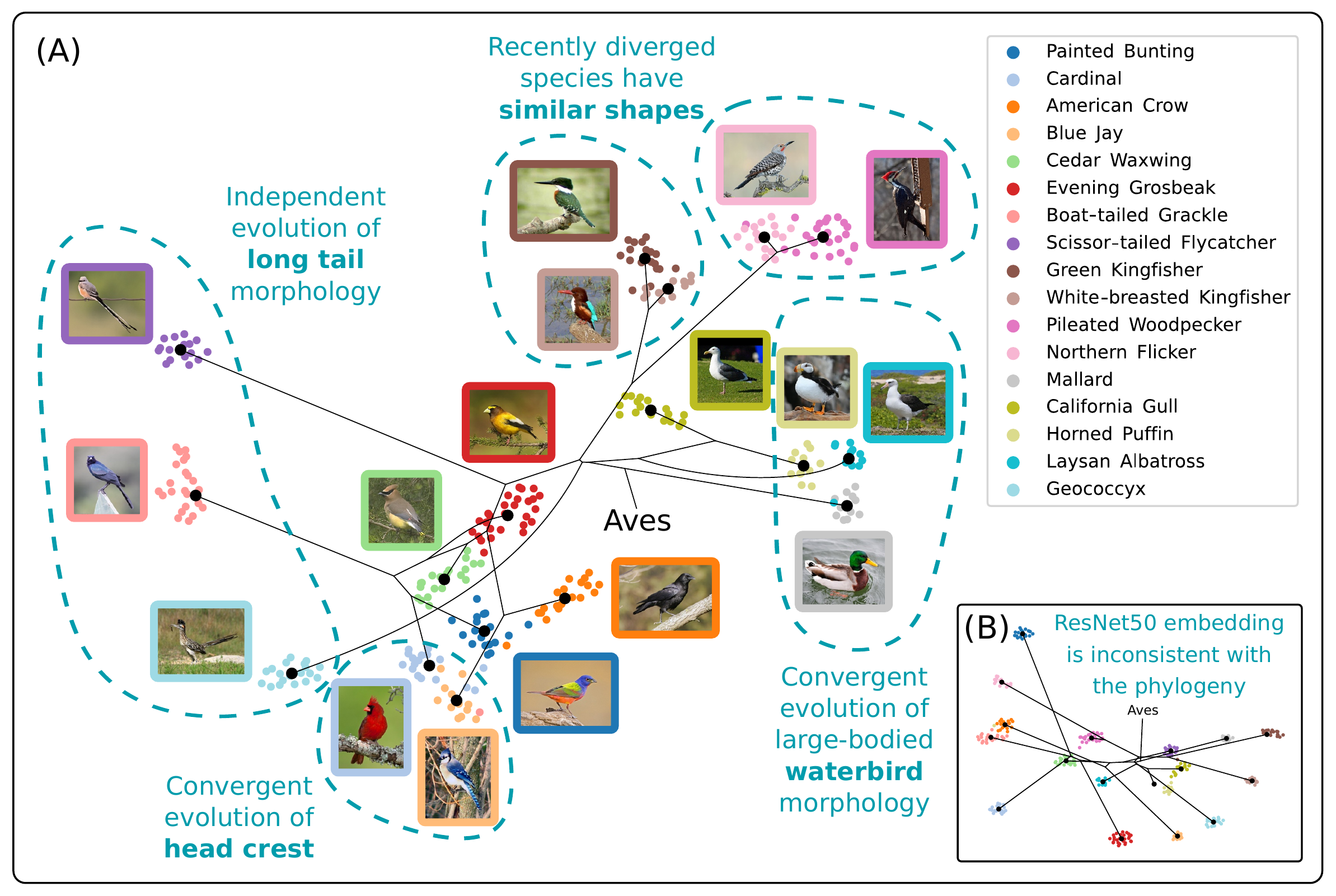}
\end{center}
   \caption{\textbf{UMAP} visualization of the principal components of the AVES shape space. (A) Different species are well separated, and similar species are embedded close to each other. Thin lines show the ancestral state reconstruction of bird shape based on the phylogeny. (B) A similar analysis demonstrates that ResNet50 features extracted from the images are inconsistent with the phylogeny.}
\label{fig:umap}
\end{figure}

We also create AVES, a new multi-species avian model by learning a PCA shape space over all species means (we observe similar expressiveness when learning over all individual reconstructions). The learned space, shown in Figure \ref{fig:ss_all}, captures characteristics across different avian species. For shape analysis (Fig. \ref{fig:ss_all},\ref{fig:umap}), the PCA space is learned with each sample normalized to have a unit body length so that the analysis is scale invariant. For all other tasks, the PCA is learned without scale normalization.

\textbf{Recovered shape variation among species is correlated with the phylogeny.} We visualize a UMAP~\cite{mcinnes2020umap} embedding of the learned shape PCA in Figure \ref{fig:umap}. Species are well separated and a phylogenetic analysis~\cite{goolsby2019rphylopars} shows that recently diverged species are embedded close together. The analysis also reveals several examples of convergent evolution for long tails, waterbird body shape, and head crests. Shape variation across species has high phylogenetic signal and is correlated with the phylogeny, or tree of relatedness~\cite{revell2010} (Figure \ref{fig:umap}, Table~\ref{tab:phylo}). On the other hand, visual features extracted using a ResNet50 embedding network trained on CUB~\cite{kim2020cubml} are not correlated with the phylogeny, despite clustering well (Table~\ref{tab:phylo}).  Avian shape captured by our shape space is therefore a more reliable phylogenetic trait than learned perceptual features. Further analyses can be found in the supplementary material.

\textbf{Comparison of species-specific models with the template model.} To evaluate whether the species-specific model can fit to unseen individuals better than the ABM template model (baseline), we split the CUB samples into 70\% training and 30\% testing for each species. We learn a shape model for each species following the main procedure on the training set. We then fit the model to the testing set by minimizing keypoint and silhouette reprojection errors (optimizing Eq~\ref{eq:obj_individual_shape} with respect to $\Theta$ and $\beta$). We evaluate two metrics: the percentage of correct keypoints (PCK) threshold at 5\% of the bounding box size, and the intersection over union (IoU) of the silhouettes. Table~\ref{tab:species-specific} compares results with the baseline. The species-specific models outperform ABM on all samples for all species, especially for species whose average shape is significantly different from the shape of the template model.

\begin{table}[bt!]
\centering
\small
\tabcolsep=0.85mm
\begin{tabular}{@{}lcc}
\toprule
Features & lambda & p-value \\
\midrule
AVES Shape PCs & $0.99 \pm 0.02$ & $<0.0001$ \\
ResNet50 & $0.18 \pm 0.20$ & 0.60 \\
\bottomrule
\end{tabular}
\vspace{-2mm}
\caption{\textbf{Captured shapes are correlated with the avian phylogeny.} A trait is consistent with a given phylogeny if it has high phylogenetic signal (Pagel's lambda~\cite{pagel1997,pagel1999,revell2010}), which measures the tendency of related species to be closer together in shape space than species drawn at random~\cite{blomberg2002signal}. Numbers are mean $\pm$ std across 100 UMAP embeddings with different initializations for both shape PCs and ResNet50 features.}
\label{tab:phylo}
\end{table}

\begin{table}
\centering
\small
\hspace{-3mm}
\tabcolsep=0.85mm

\begin{tabular}{@{}lcccccc } 
\toprule
 & \multicolumn{2}{c}{Species A} 
 & \multicolumn{2}{c}{Species B} 
 & \multicolumn{2}{c}{All 17 species} \\
\cmidrule(lr){2-3} \cmidrule(lr){4-5} \cmidrule(lr){6-7}
& PCK05 & IoU & PCK05 & IoU & PCK05 & IoU \\ 
\midrule
Baseline (ABM) & 0.941 & 0.799 & 0.809 & 0.723 & 0.857 & 0.765 \\ 
Species-specific & 0.988 & 0.816 & 0.963 & 0.782 & \textbf{0.954} & \textbf{0.805} \\
\bottomrule
\end{tabular}

\vspace{-2mm}
\caption{\textbf{Species-specific models} are fitted to the test set of each species. Species A is the painted bunting, which is similar in shape to baseline template. We see a larger performance gain for species that have a very different shape from the template, such as species B, the boat-tailed grackle.}
\label{tab:species-specific}
\end{table}

\begin{table}
\centering
\small
\hspace{-3mm}
\tabcolsep=0.85mm

\begin{tabular}{@{}lcccccc } 
\toprule
 & \multicolumn{2}{c}{Species A} 
 & \multicolumn{2}{c}{Species B} 
 & \multicolumn{2}{c}{All 17 species} \\
\cmidrule(lr){2-3} \cmidrule(lr){4-5} \cmidrule(lr){6-7}
& PCK05 & IoU & PCK05 & IoU & PCK05 & IoU \\ 
\midrule
Baseline (ABM) & 0.940 & 0.784 & 0.819 & 0.723 & 0.857 & 0.756 \\ 
Multi-species & 0.969 & 0.813 & 0.964 & 0.793 & \textbf{0.963} & \textbf{0.804} \\
\bottomrule
\end{tabular}

\vspace{-2mm}
\caption{\textbf{Multi-species model} learned on $(k-1)$ species and test on the held-out species via $k$-fold cross-validation. Species A and B are the same as in Table \ref{tab:species-specific}.}
\label{tab:multi-species}
\vspace{-2mm}
\end{table}

\textbf{Generalization capability of the multi-species model.} To evaluate whether the AVES model can generalize to new species, we conduct $k$-fold cross-validation where $k=17$ is the number of species. Every time we learn a new version of AVES on $(k-1)$ species and test it on the held-out species. We fit the model to the held-out species by minimizing the 2D reprojection errors similarly to testing the species-specific models. Table~\ref{tab:multi-species} summarizes the results, and shows that the AVES model can indeed generalize to unseen species.

\textbf{Evaluation on the Aviary dataset.}
The ABM model was introduced to reconstruct the brown-headed cowbirds in the Penn Aviary dataset~\cite{badger20203d}, which provides multi-view annotations of bird instances across several temporal slices.

We conduct an experiment to see whether the multi-species model can improve over the baseline results. We initialize the shape coefficients to one that matches the default shape. We then fixed the coefficient and fit the model to multi-view samples following the baseline procedure described in \cite{badger20203d}. To get our results, we allow the shape coefficient to vary towards the end of the optimization. We compare the results in Table~\ref{tab:quantresults}.

Although the default shape is already a good approximation for cowbirds, we observe improvement in PCKs and comparable results in IoU, indicating that our new shape model AVES better estimates the birds' shapes. We provide additional 3D evaluations of the main approach in the supplementary material.

\begin{table}
\centering
\small
\hspace{-3mm}
\tabcolsep=0.85mm
\begin{tabular}{@{}lccc}
\toprule
Method & PCK05 & PCK10 & IoU \\
\midrule
Cowbird~\cite{badger20203d} & 0.406 & 0.723 & 0.605 \\
Ours (AVES) & \textbf{0.432} & \textbf{0.742} & \textbf{0.606} \\
\midrule
Cowbird~\cite{badger20203d} w/ silhouette & 0.412 & 0.731 & 0.631 \\
Ours (AVES) w/ silhouette & \textbf{0.429} & \textbf{0.742} & \textbf{0.632} \\
\bottomrule
\end{tabular}
\vspace{-2mm}
\caption{\textbf{Quantitative evaluation on the Aviary dataset.} We deploy the new shape model on the Aviary dataset to reconstruct examples from multiple views, and compare results to the baseline.}
\label{tab:quantresults}
\vspace{-2mm}
\end{table}

\begin{table}
\centering
\small
\hspace{-3mm}
\tabcolsep=0.85mm
\begin{tabular}{@{}lccc}
\toprule
Method & PCK05 & PCK10 & IoU \\
\midrule
CMR~\cite{kanazawa2018learning} & 0.432 & 0.811 & 0.703 \\
Baseline (ABM)~\cite{badger20203d} & 0.679 & 0.923 & 0.706 \\
Ours (AVES) & \textbf{0.703} & \textbf{0.931} & \textbf{0.720} \\
\bottomrule
\end{tabular}
\vspace{-2mm}
\caption{\textbf{Quantitative evaluation of different regression methods on the CUB dataset.} We compare the network that predicts the AVES model parameters with the baseline that predicts the ABM parameters and with CMR.}
\label{tab:regression}
\vspace{-2mm}
\end{table}

\begin{figure*}[!t]
	\centering
    \includegraphics[width=0.10\textwidth]{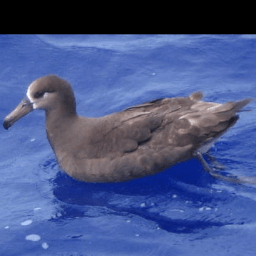}
    \includegraphics[width=0.10\textwidth]{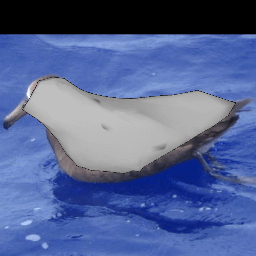}
    \includegraphics[width=0.10\textwidth]{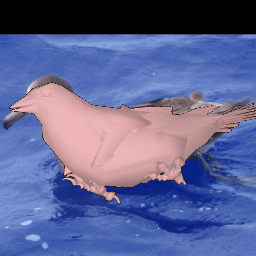}
    \includegraphics[width=0.10\textwidth]{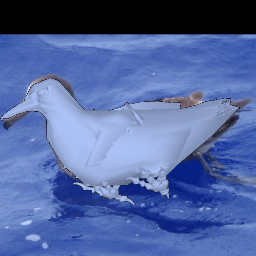}
     \includegraphics[width=0.10\textwidth]{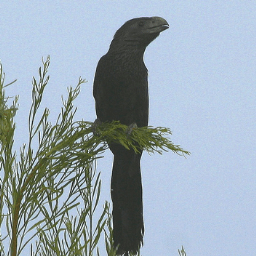}
    \includegraphics[width=0.10\textwidth]{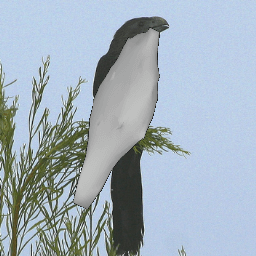}
    \includegraphics[width=0.10\textwidth]{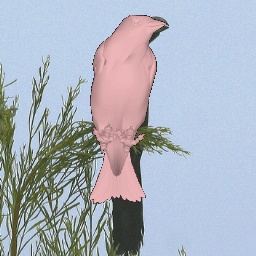}
    \includegraphics[width=0.10\textwidth]{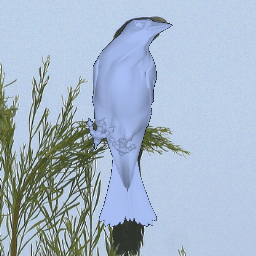}\\
     \includegraphics[width=0.10\textwidth]{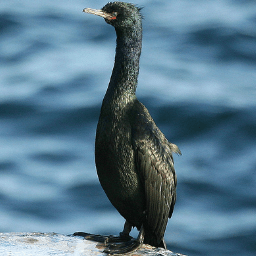}
    \includegraphics[width=0.10\textwidth]{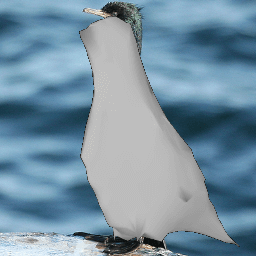}
    \includegraphics[width=0.10\textwidth]{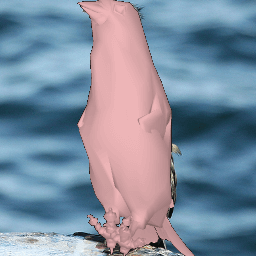}
    \includegraphics[width=0.10\textwidth]{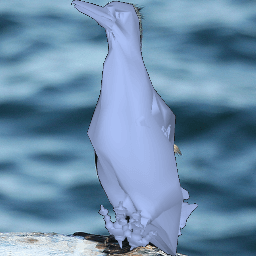}
     \includegraphics[width=0.10\textwidth]{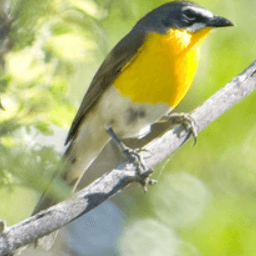}
    \includegraphics[width=0.10\textwidth]{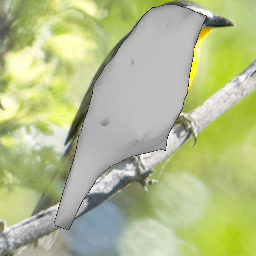}
    \includegraphics[width=0.10\textwidth]{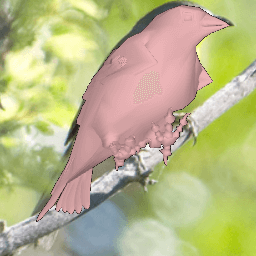}
    \includegraphics[width=0.10\textwidth]{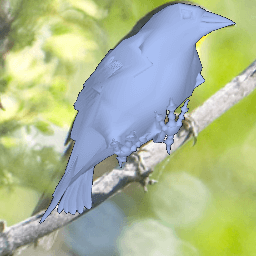}\\
	\vspace{-2mm}
	\caption{{\bf Qualitative comparison of regression-based methods.}
Gray: Reconstruction by CMR \cite{kanazawa2018learning}. Pink: Baseline (ABM) \cite{badger20203d}. Blue: Ours (AVES).
More qualitative results can be found in the Sup.Mat.
}
\label{fig:regression}
\vspace{-2mm}
\end{figure*}

\textbf{Regressing 3D bird shape from a single RGB image.} To demonstrate the effectiveness of the AVES model, we train a neural network similar to HMR~\cite{kanazawa2018end} that regresses the AVES model parameters from a single image. The network is supervised using 2D keypoint and silhouette reprojection losses. We compare our approach against a baseline that regresses the ABM~\cite{badger20203d} parameters and with CMR~\cite{kanazawa2018end}. Following  CMR, we split the test set of CUB into a validation set used for hyperparameter tuning and a separate test set. For more details about the training and the model we refer the reader to the supplementary material. In Figure~\ref{fig:regression} we show qualitative comparisons. Using a parametric bird model improves the 3D shape over the model-free approach and in turn AVES is able to better capture shape variation across different bird species. Table~\ref{tab:regression} presents a quantitative comparisons using the available 2D annotations. The AVES model clearly outperforms both the ABM model and CMR on the standard 2D benchmarks.

\begin{figure}[]
    \center
	\vspace{-2mm}
\includegraphics[width=0.49\linewidth]{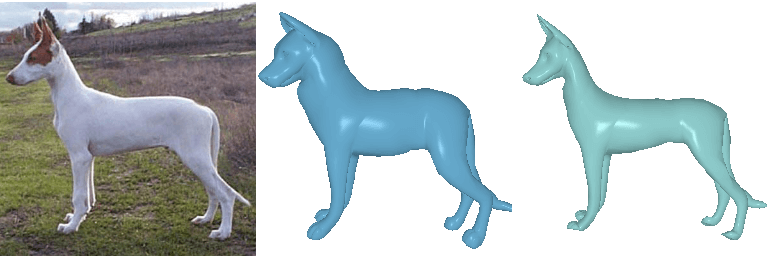}
\includegraphics[width=0.49\linewidth]{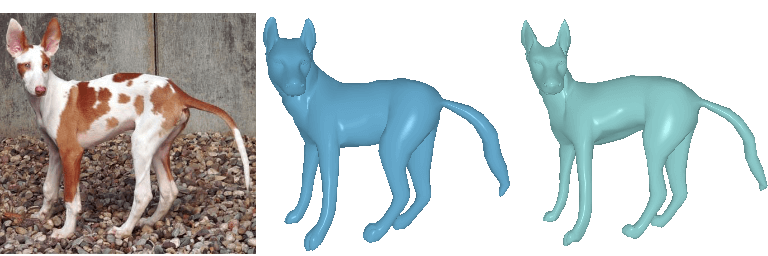} \\
\includegraphics[width=0.49\linewidth]{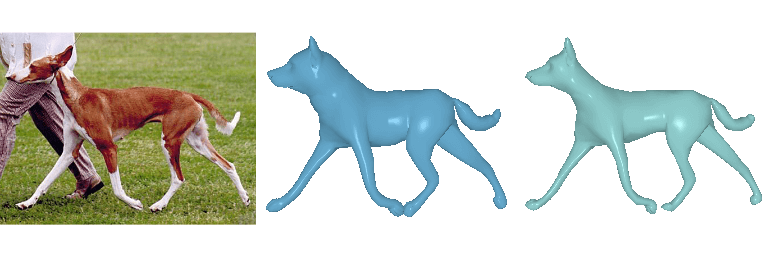}
\includegraphics[width=0.49\linewidth]{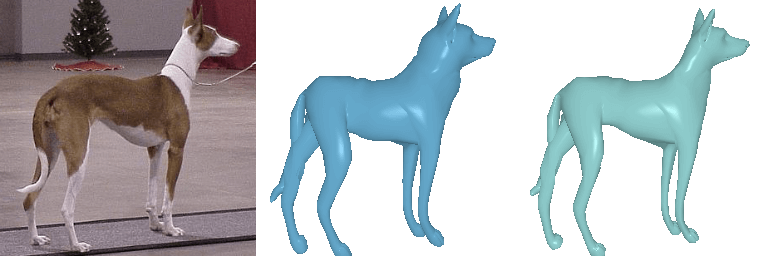}
\vspace{-2mm}
\caption{\textbf{Application in other animals.} From left to right are images, alignments with SMAL Canis, and our results.
}
\label{fig:dog}
\vspace{-2mm}
\end{figure}

\textbf{Application to other animals}. Our method is general and can be used to capture other types of animals. We apply it to dogs, which have articulations and intra-breed shape variation, similar to the scenario examined for birds.
We use the mean Canis shape from SMAL~\cite{zuffi20173d} and the limb scaling implementation from SMBLD~\cite{biggs2020left}. Different than previous works, we do not utilize the shape space. We optimize the pose and limb scales to align the template to images, shown as dark blue in Fig.~\ref{fig:dog}. Given the estimated alignments, we follow our method to capture the shape variation. Fig.~\ref{fig:dog} shows qualitative results of the Ibizan hounds from Stanford Dogs~\cite{biggs2020left, khosla2011novel}. 

\textbf{Limitations and failures}. 
We propose a method to capture animal shape space with an articulated template. The quality of the recovered shapes is largely influenced by the accuracy of the estimated pose. Our assumption is that the template can be aligned accurately with the images so that the shape difference is inferred through silhouette difference. Achieving accurate alignment could be difficult for many reasons. First, the pose might not be represented in the pose prior. In our case, the pose of flying birds or lying dogs cannot be solved reliably. Secondly, alignment is ill-defined if the target species has a very different body proportion than the template. The bone parameter of the bird model and the limb scaling of the dog model allow body proportion to adapt and therefore improve alignments. We include some failure cases in the Supplementary Material.

\section{Conclusion}
In this work, we present a new method for capturing shape models of novel species from image sets. Our method starts from an articulated mesh model and learns intra-species and inter-species deformations using only 2D annotations. Using the captured species-specific shapes, we develop a multi-species statistical shape model, AVES, which is correlated with the avian phylogeny and accurately reconstructs individuals. We use our AVES model for various reconstruction tasks such as model fitting and 3D shape regression from a single RGB image. Our approach focuses on birds but can be applied to other animal classes. Because we use pose priors and predefined kinematic chain, our models cannot capture some extreme pose and shape changes. We leave this challenge for future work.

\smallbreak
\begin{small}
\noindent\textbf{Acknowledgements:} We gratefully appreciate support through the following grants: NSF IIS 1703319, NSF MRI 1626008, NSF TRIPODS 1934960, and NSF CPS 2038873.
\end{small}

\begin{figure*}[!t]
	\centering
    \includegraphics[width=0.10\textwidth]{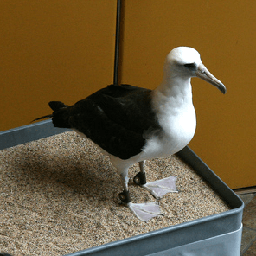}
    \includegraphics[width=0.10\textwidth]{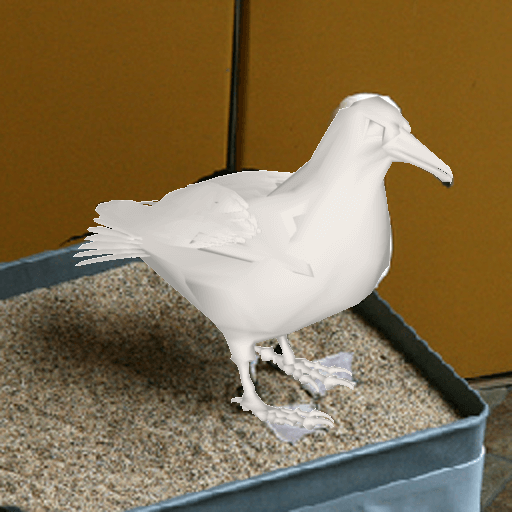}
    \includegraphics[width=0.10\textwidth]{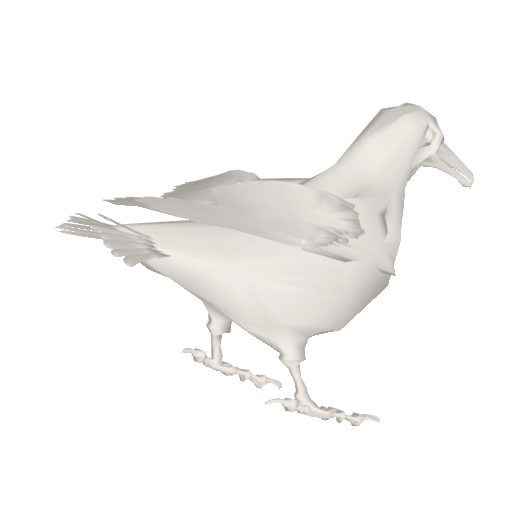}
    \includegraphics[width=0.10\textwidth]{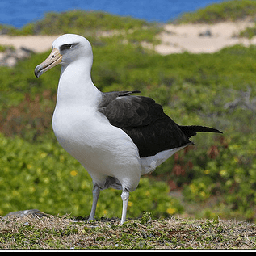}
    \includegraphics[width=0.10\textwidth]{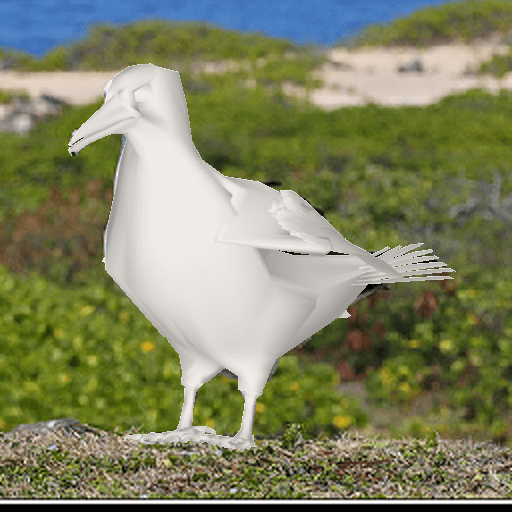}
    \includegraphics[width=0.10\textwidth]{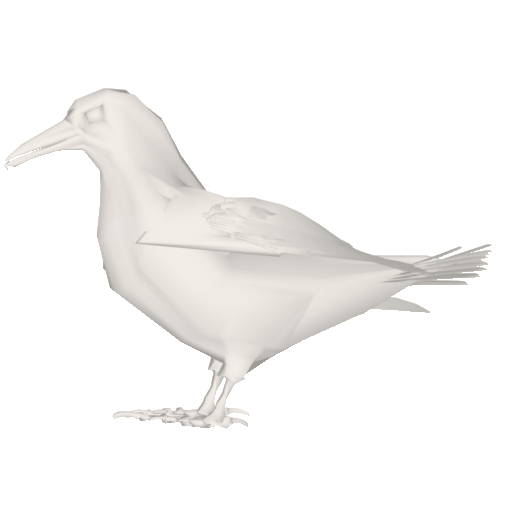}
    \includegraphics[width=0.10\textwidth]{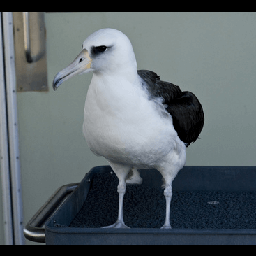}
    \includegraphics[width=0.10\textwidth]{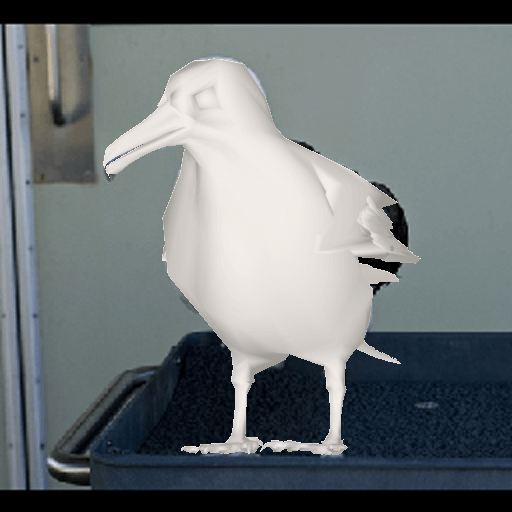}
    \includegraphics[width=0.10\textwidth]{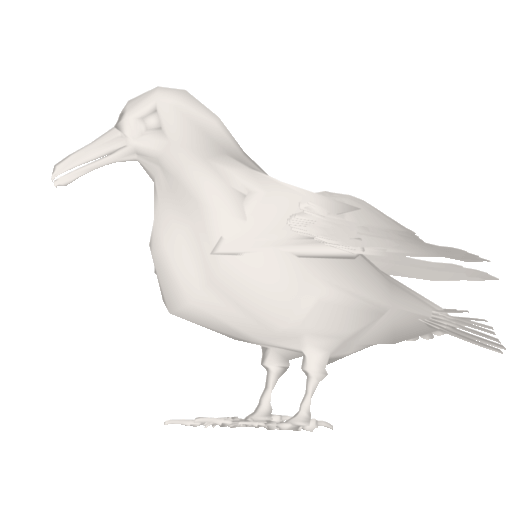} \\
    \includegraphics[width=0.10\textwidth]{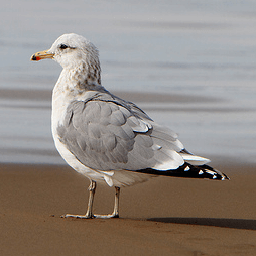}
    \includegraphics[width=0.10\textwidth]{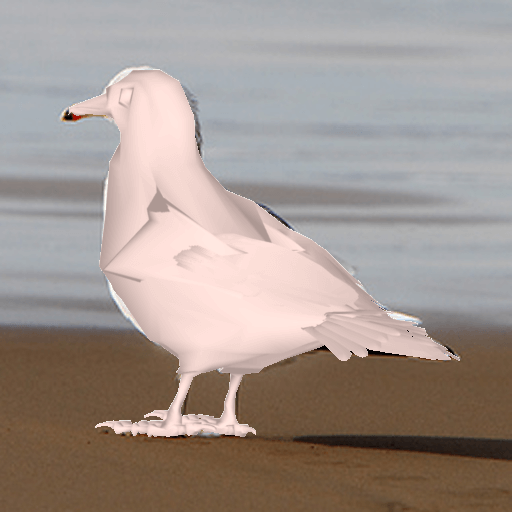}
    \includegraphics[width=0.10\textwidth]{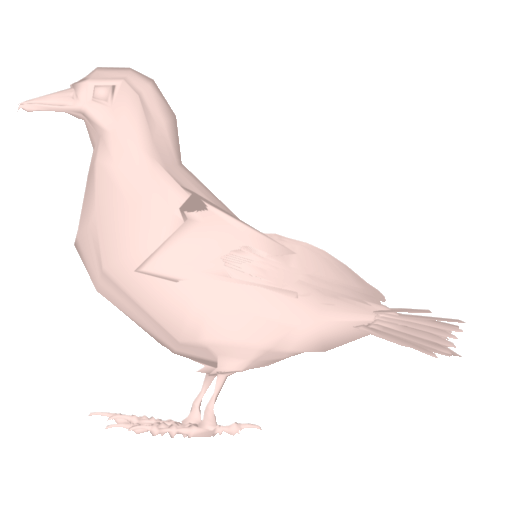}
    \includegraphics[width=0.10\textwidth]{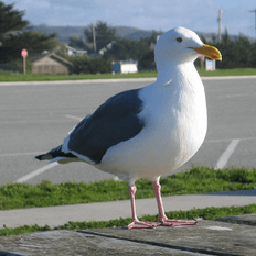}
    \includegraphics[width=0.10\textwidth]{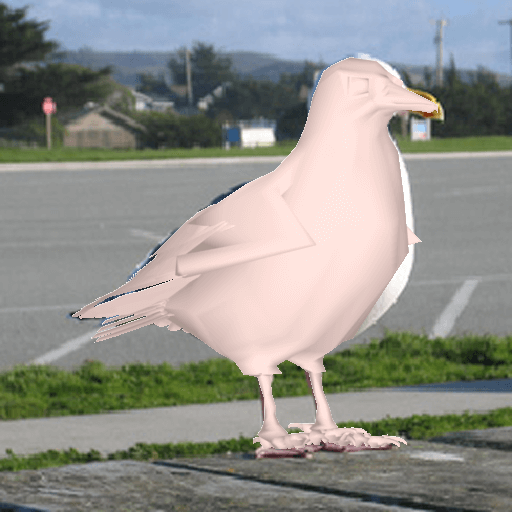}
    \includegraphics[width=0.10\textwidth]{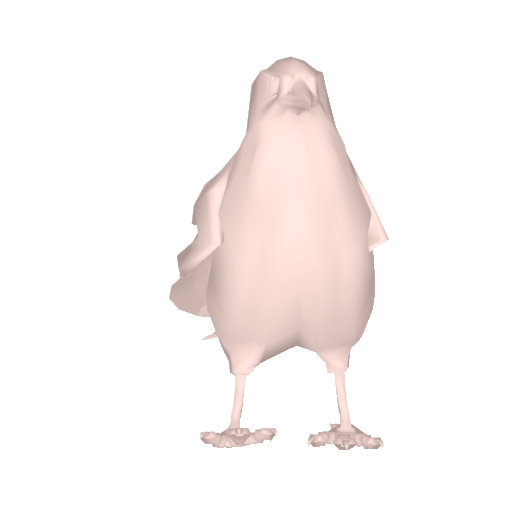}
    \includegraphics[width=0.10\textwidth]{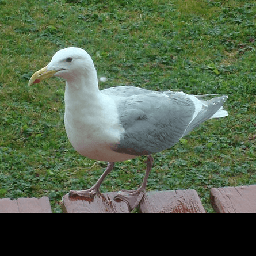}
    \includegraphics[width=0.10\textwidth]{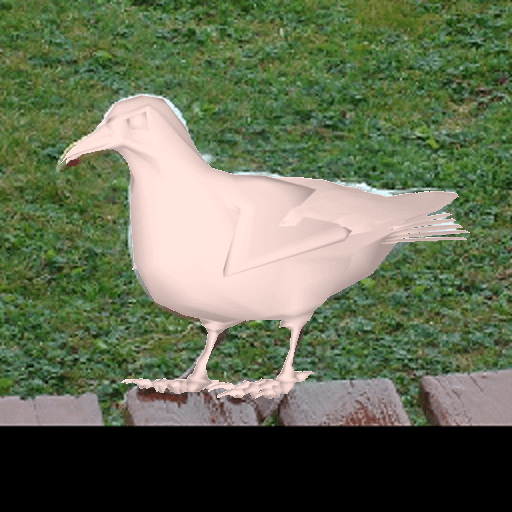}
    \includegraphics[width=0.10\textwidth]{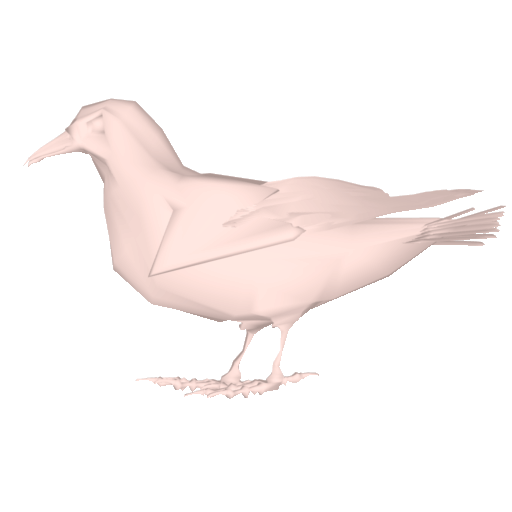}\\
    \includegraphics[width=0.10\textwidth]{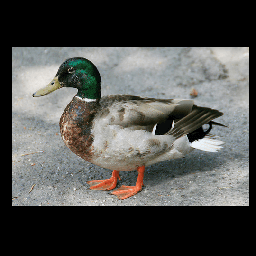}
    \includegraphics[width=0.10\textwidth]{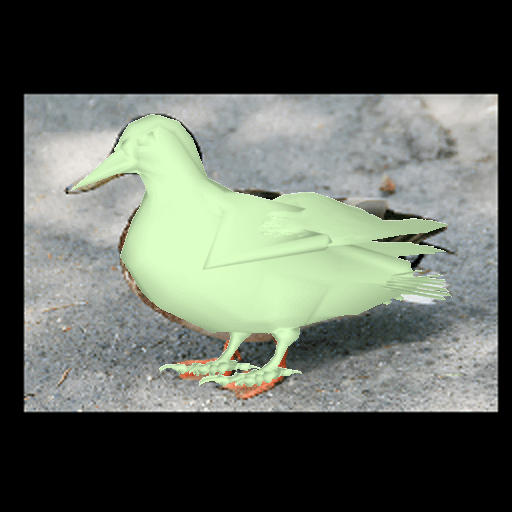}
    \includegraphics[width=0.10\textwidth]{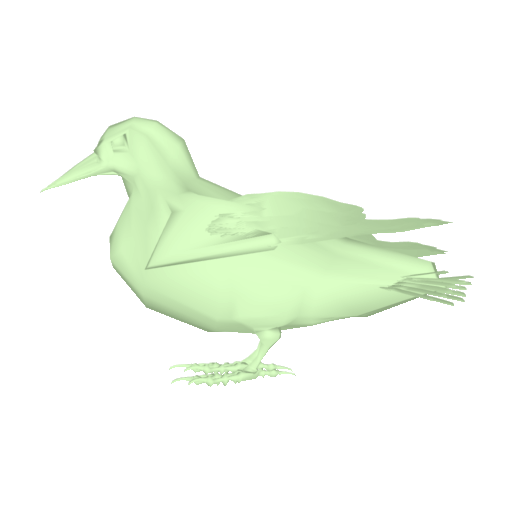}
    \includegraphics[width=0.10\textwidth]{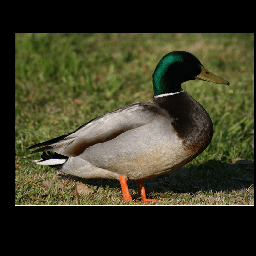}
    \includegraphics[width=0.10\textwidth]{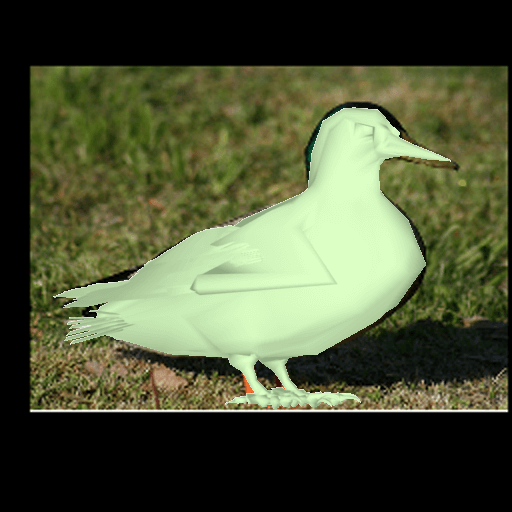}
    \includegraphics[width=0.10\textwidth]{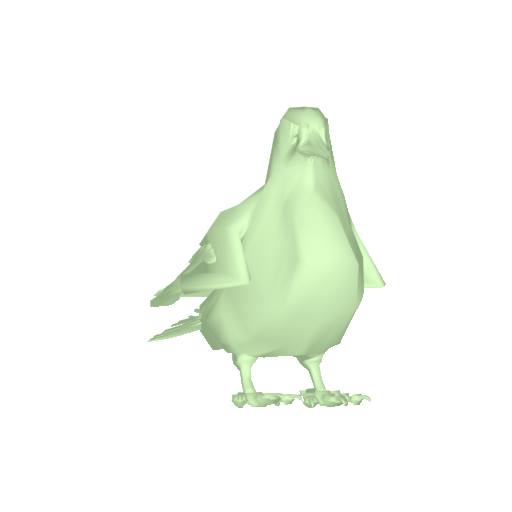}
    \includegraphics[width=0.10\textwidth]{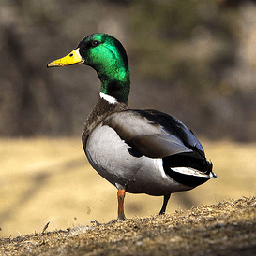}
    \includegraphics[width=0.10\textwidth]{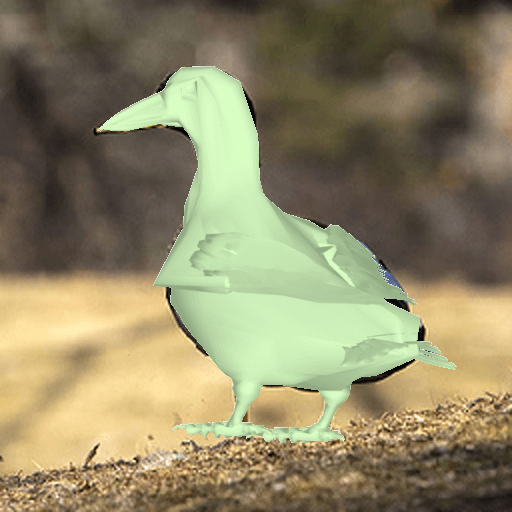}
    \includegraphics[width=0.10\textwidth]{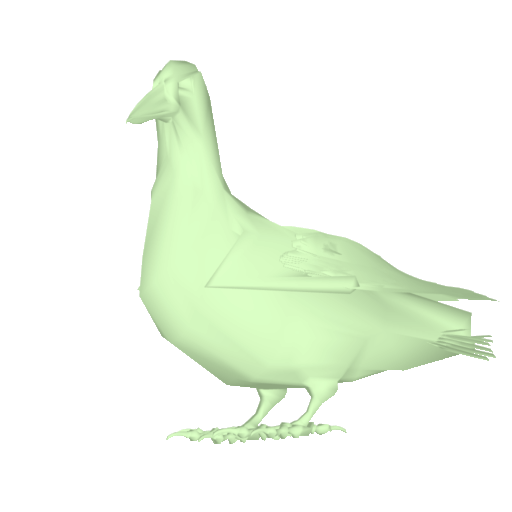}\\
    \includegraphics[width=0.10\textwidth]{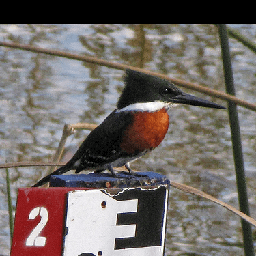}
    \includegraphics[width=0.10\textwidth]{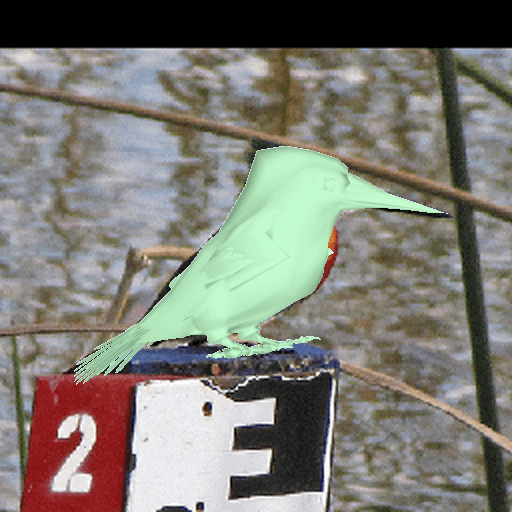}
    \includegraphics[width=0.10\textwidth]{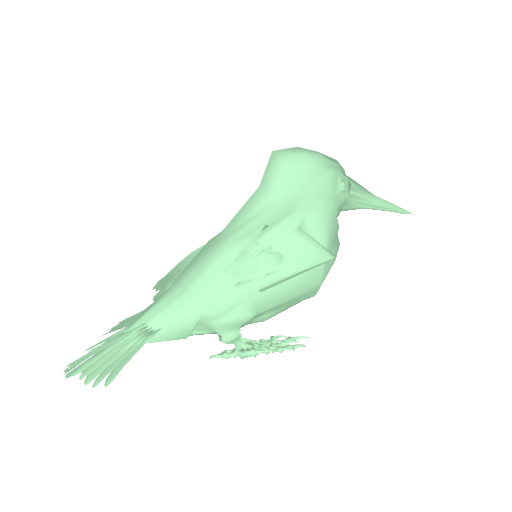}
    \includegraphics[width=0.10\textwidth]{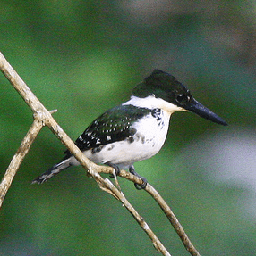}
    \includegraphics[width=0.10\textwidth]{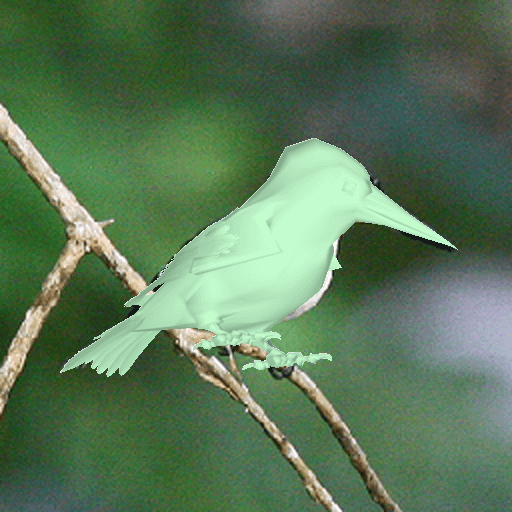}
    \includegraphics[width=0.10\textwidth]{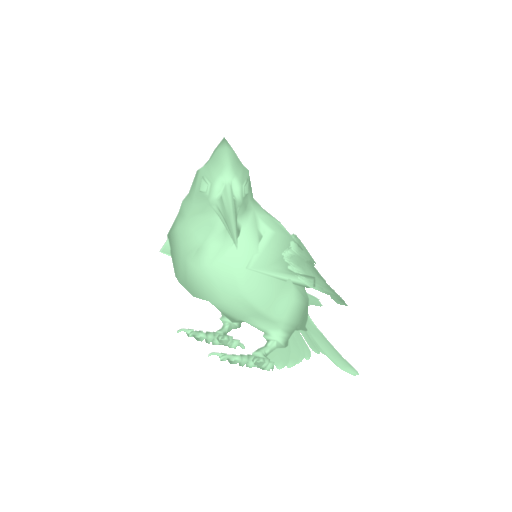}
    \includegraphics[width=0.10\textwidth]{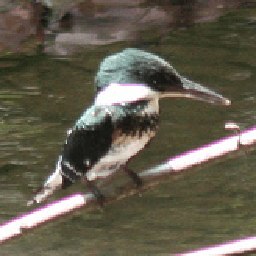}
    \includegraphics[width=0.10\textwidth]{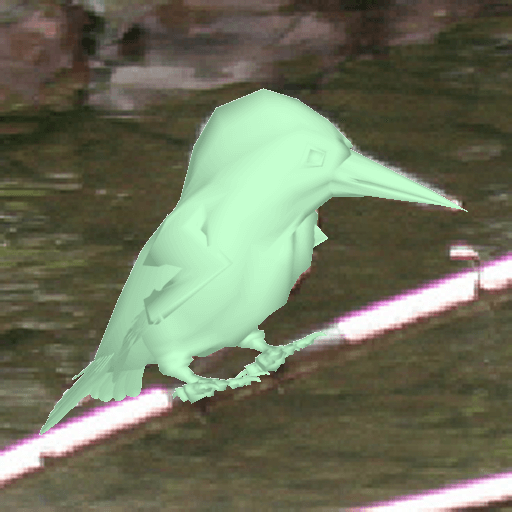}
    \includegraphics[width=0.10\textwidth]{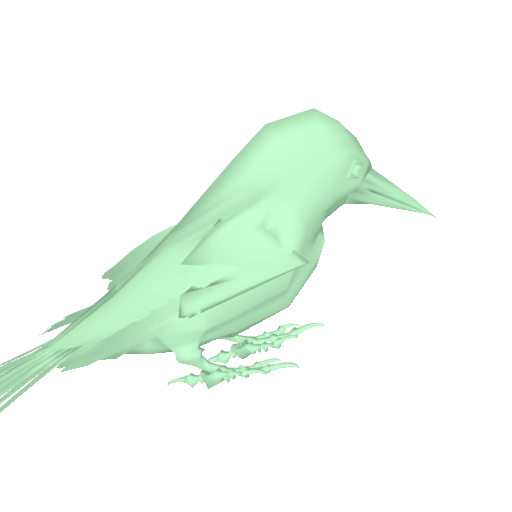}\\
    \includegraphics[width=0.10\textwidth]{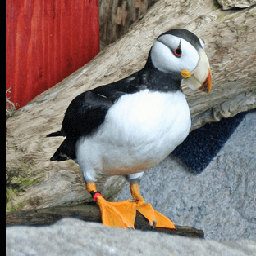}
    \includegraphics[width=0.10\textwidth]{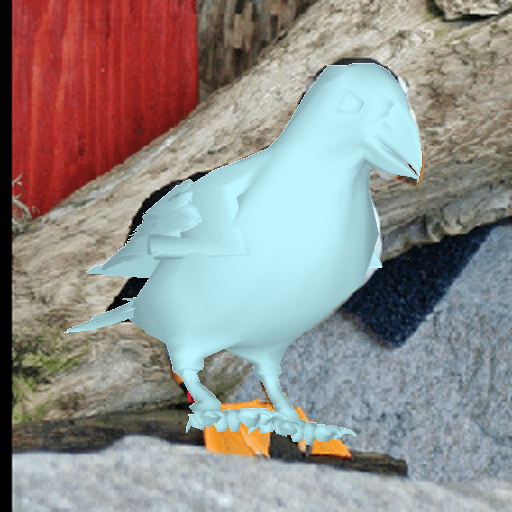}
    \includegraphics[width=0.10\textwidth]{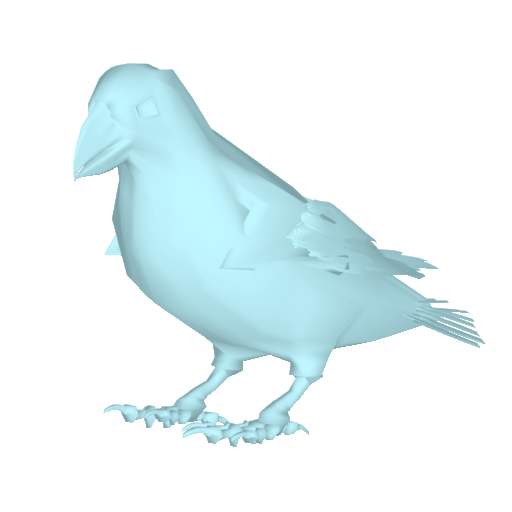}
    \includegraphics[width=0.10\textwidth]{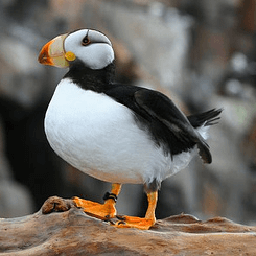}
    \includegraphics[width=0.10\textwidth]{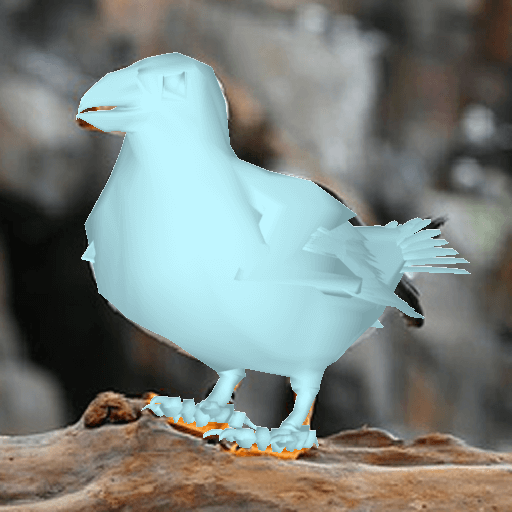}
    \includegraphics[width=0.10\textwidth]{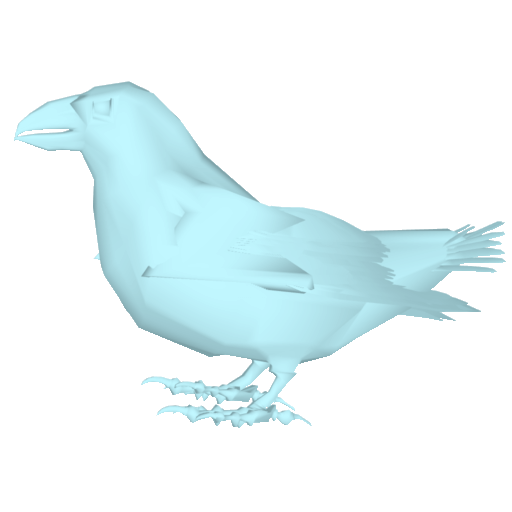}
    \includegraphics[width=0.10\textwidth]{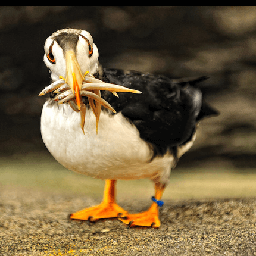}
    \includegraphics[width=0.10\textwidth]{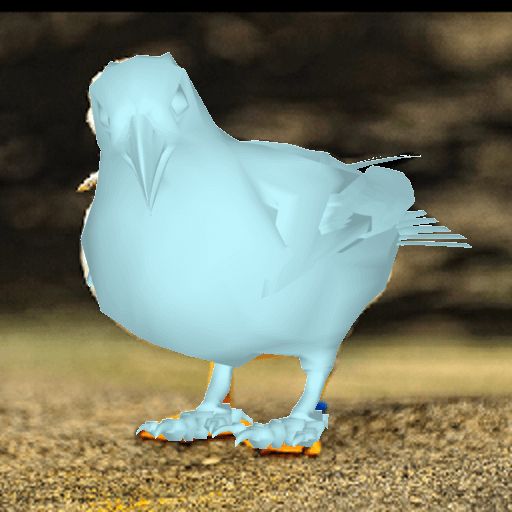}
    \includegraphics[width=0.10\textwidth]{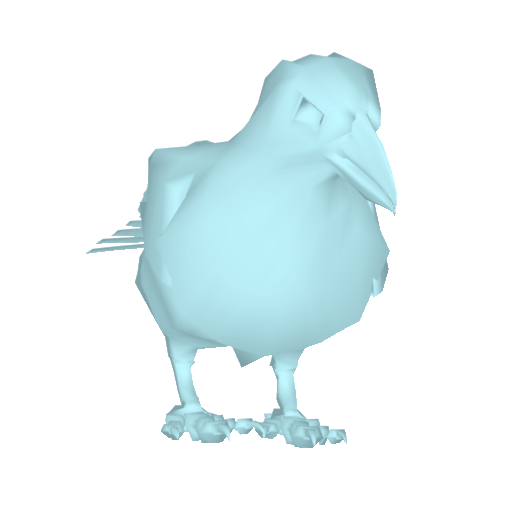}\\
    \includegraphics[width=0.10\textwidth]{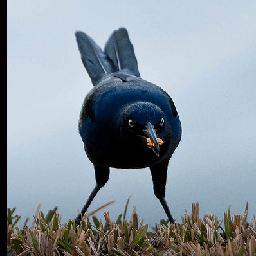}
    \includegraphics[width=0.10\textwidth]{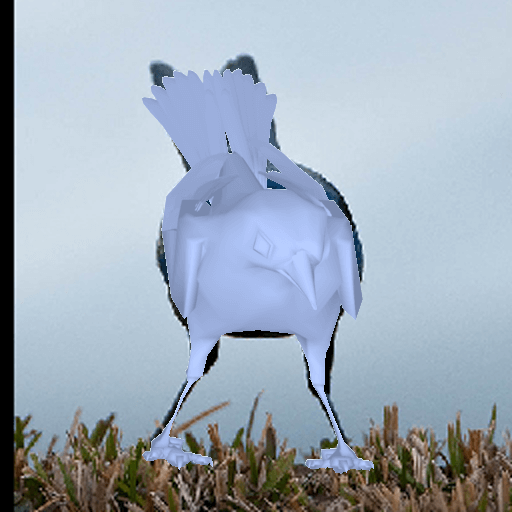}
    \includegraphics[width=0.10\textwidth]{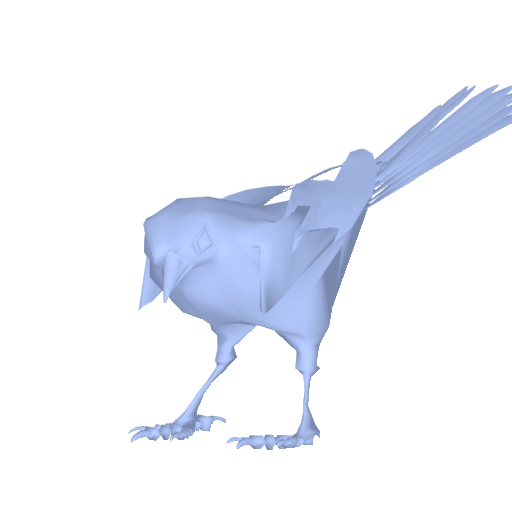}
    \includegraphics[width=0.10\textwidth]{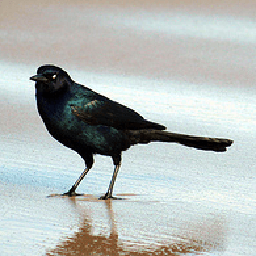}
    \includegraphics[width=0.10\textwidth]{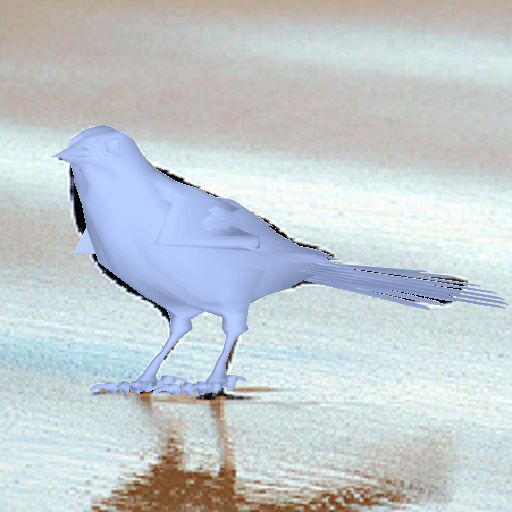}
    \includegraphics[width=0.10\textwidth]{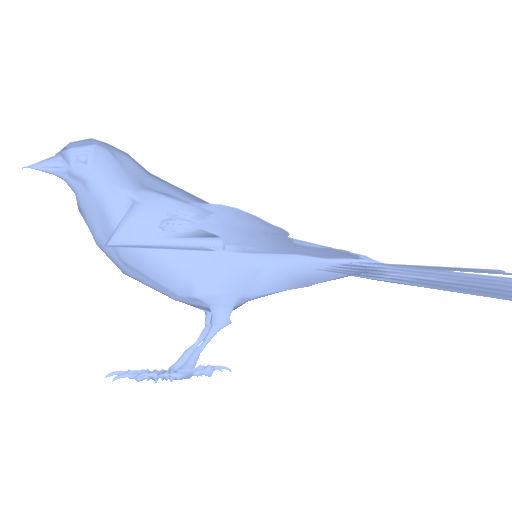}
    \includegraphics[width=0.10\textwidth]{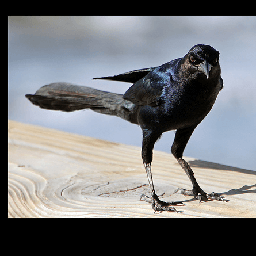}
    \includegraphics[width=0.10\textwidth]{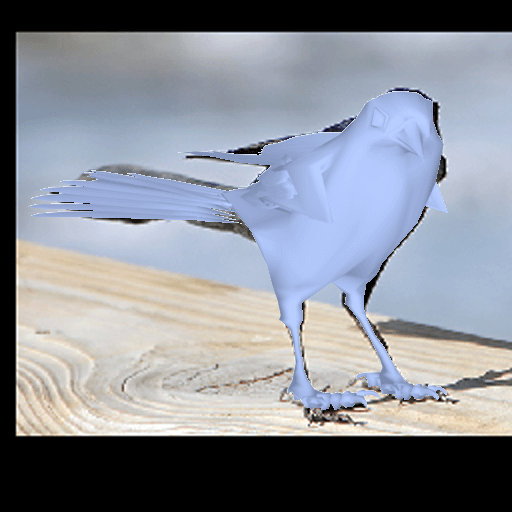}
    \includegraphics[width=0.10\textwidth]{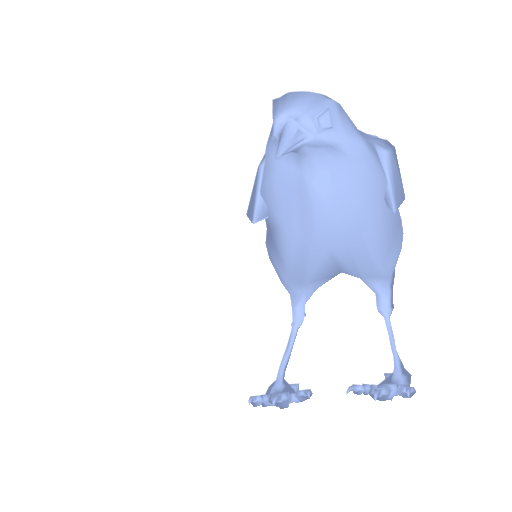}\\
    \includegraphics[width=0.10\textwidth]{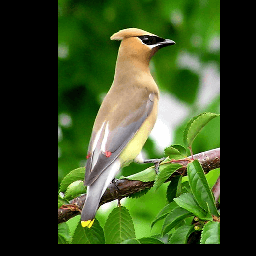}
    \includegraphics[width=0.10\textwidth]{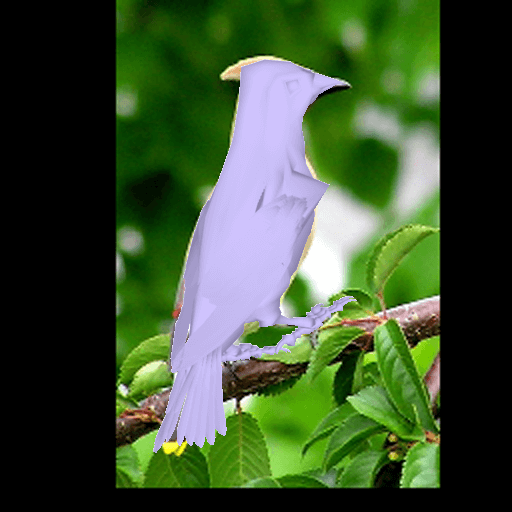}
    \includegraphics[width=0.10\textwidth]{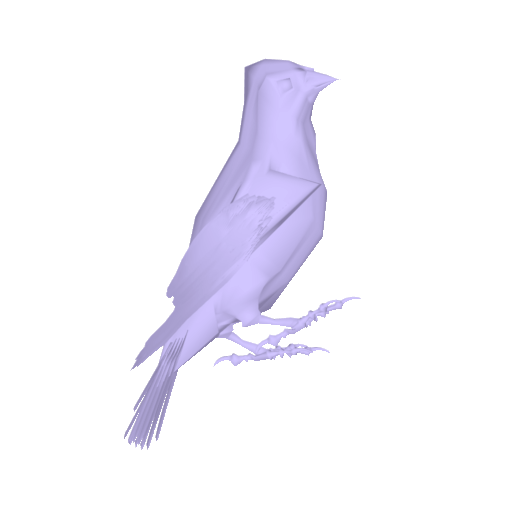}
    \includegraphics[width=0.10\textwidth]{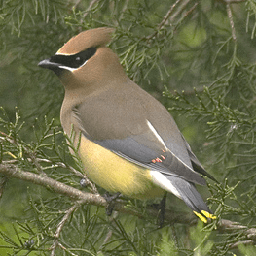}
    \includegraphics[width=0.10\textwidth]{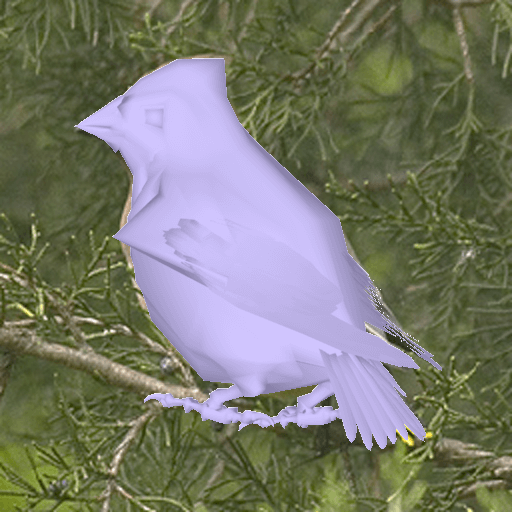}
    \includegraphics[width=0.10\textwidth]{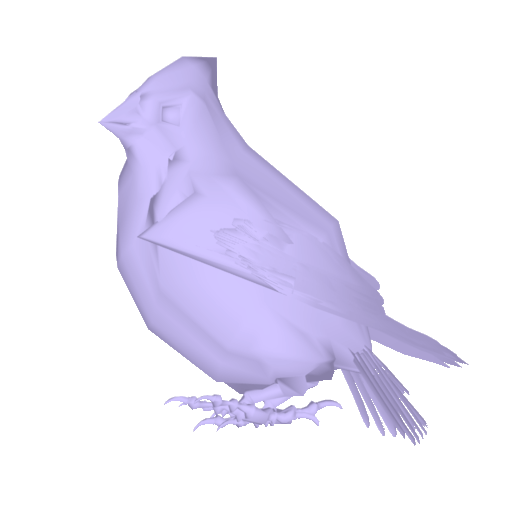}
    \includegraphics[width=0.10\textwidth]{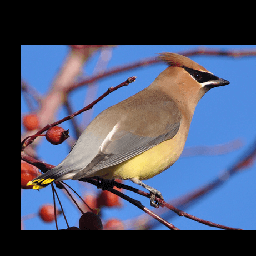}
    \includegraphics[width=0.10\textwidth]{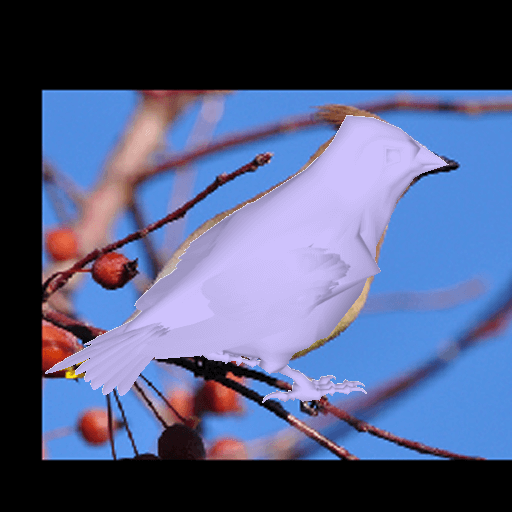}
    \includegraphics[width=0.10\textwidth]{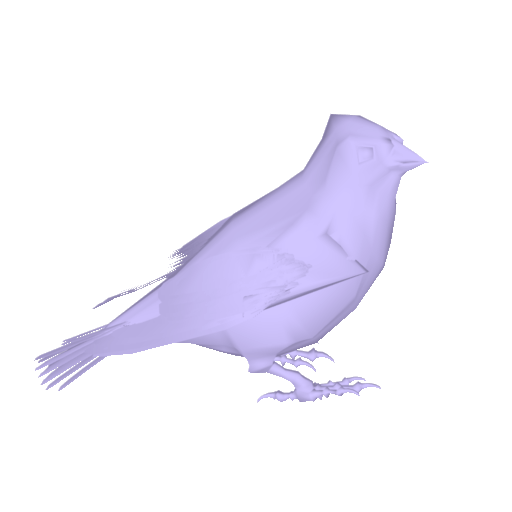}\\
    \includegraphics[width=0.10\textwidth]{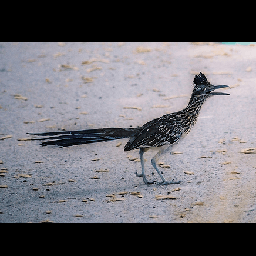}
    \includegraphics[width=0.10\textwidth]{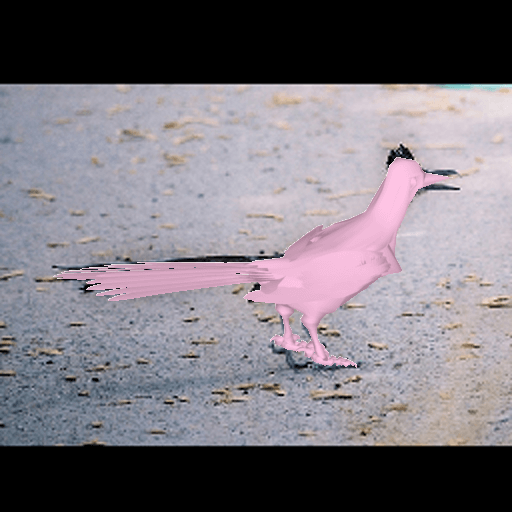}
    \includegraphics[width=0.10\textwidth]{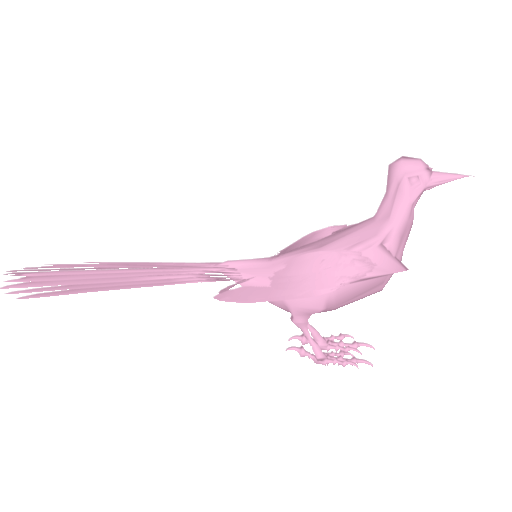}
    \includegraphics[width=0.10\textwidth]{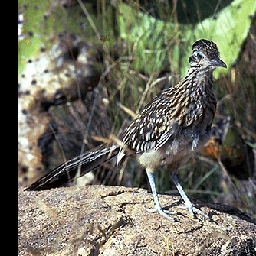}
    \includegraphics[width=0.10\textwidth]{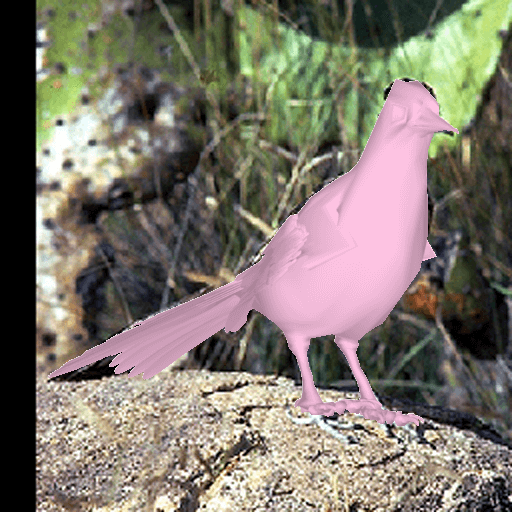}
    \includegraphics[width=0.10\textwidth]{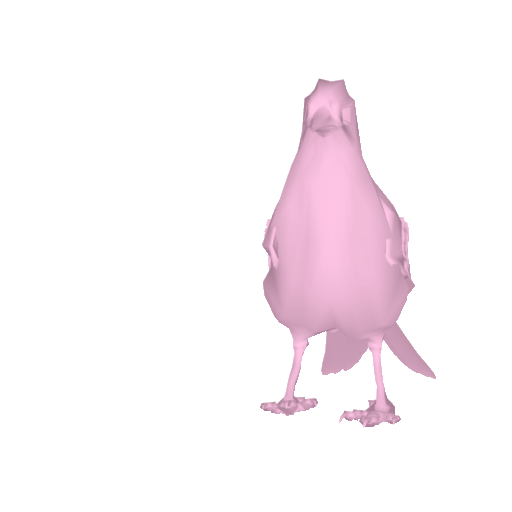}
    \includegraphics[width=0.10\textwidth]{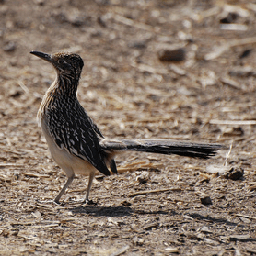}
    \includegraphics[width=0.10\textwidth]{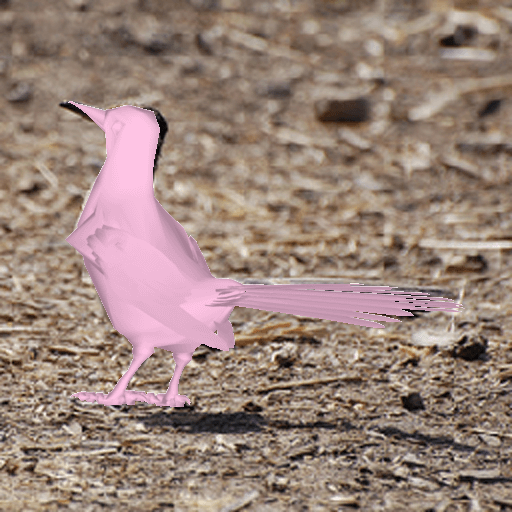}
    \includegraphics[width=0.10\textwidth]{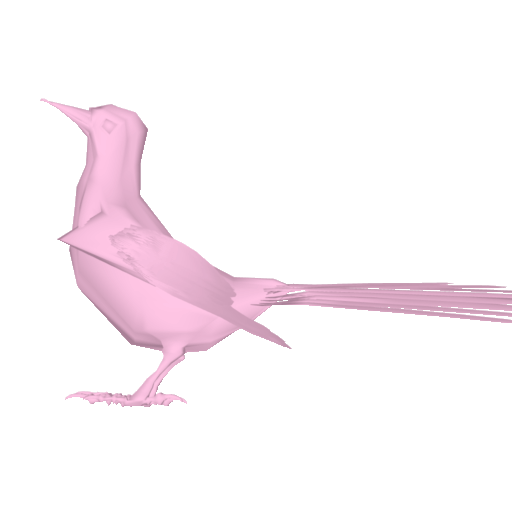}\\
    \includegraphics[width=0.10\textwidth]{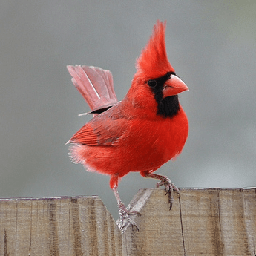}
    \includegraphics[width=0.10\textwidth]{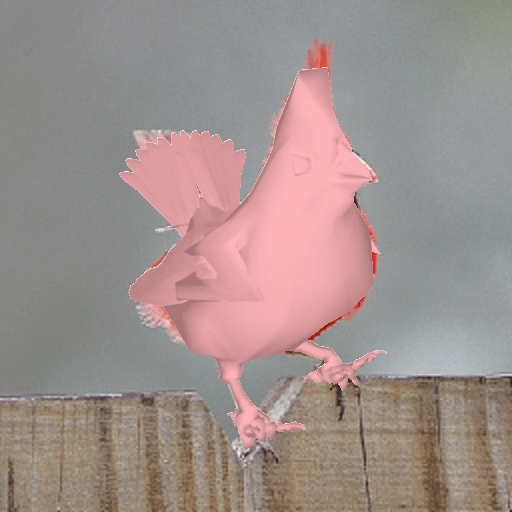}
    \includegraphics[width=0.10\textwidth]{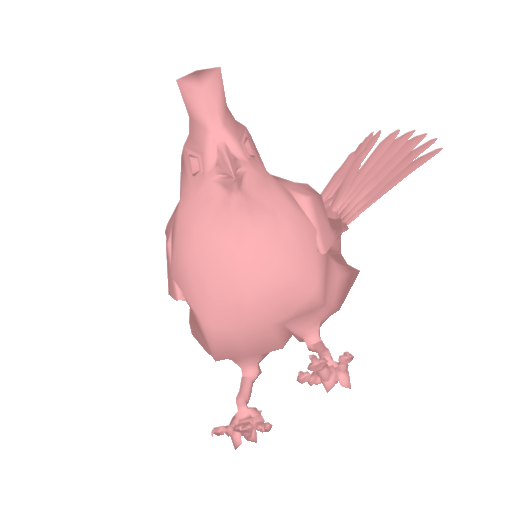}
    \includegraphics[width=0.10\textwidth]{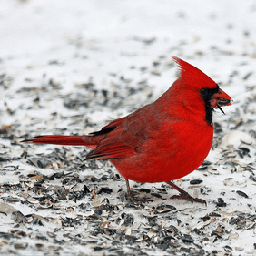}
    \includegraphics[width=0.10\textwidth]{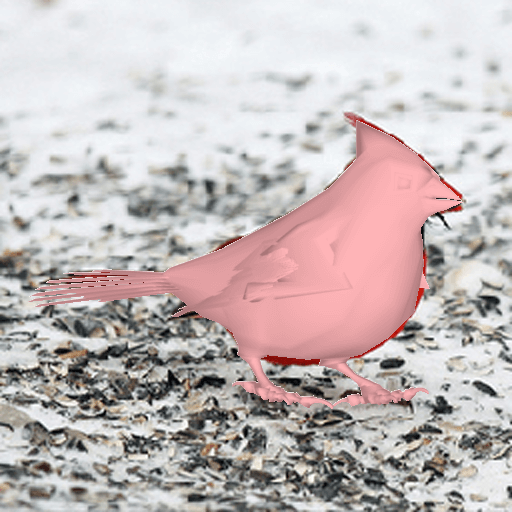}
    \includegraphics[width=0.10\textwidth]{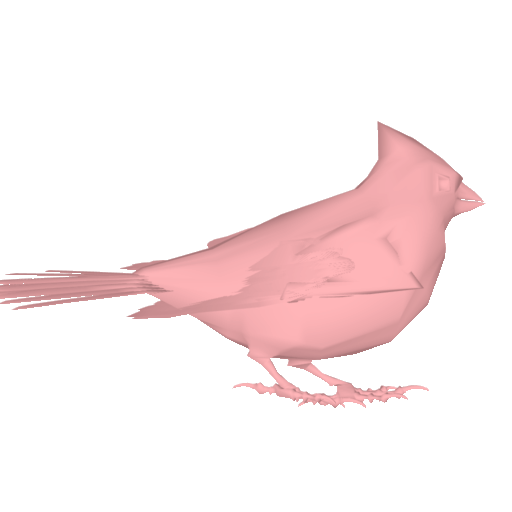}
    \includegraphics[width=0.10\textwidth]{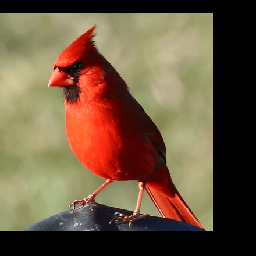}
    \includegraphics[width=0.10\textwidth]{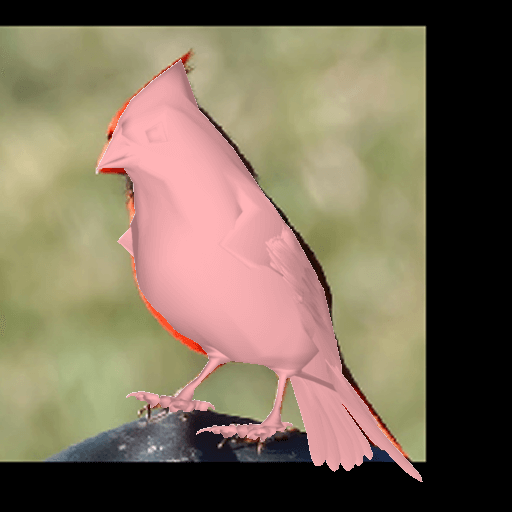}
    \includegraphics[width=0.10\textwidth]{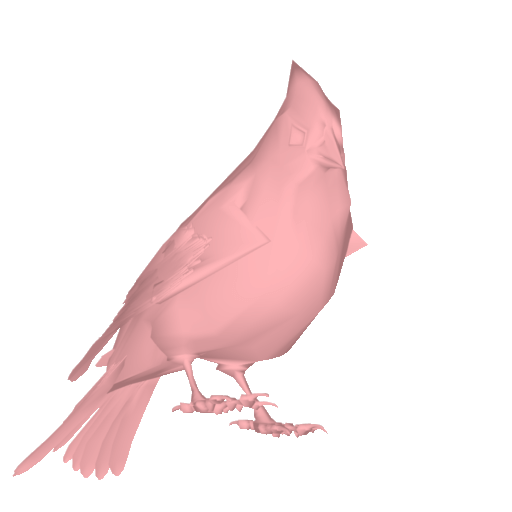}\\
	\vspace{-2mm}
	\caption{{\bf Examples of learnt species-specific models}. Each row depicts reconstructions using a particular species-specific model that includes identity-specific deformations. Each triplet includes the input image, the reconstructed mesh and the reconstructed mesh from a novel viewpoint.}
\label{fig:reconstruction}
\vspace{-2mm}
\end{figure*}

\clearpage 
{\small
\bibliographystyle{ieee_fullname}
\bibliography{egbib}
}

\end{document}